\pdfoutput=1

\documentclass[11pt]{article}

\usepackage{acl}

\usepackage{times}
\usepackage{latexsym}

\usepackage[T1]{fontenc}

\usepackage[utf8]{inputenc}

\usepackage{microtype}

\usepackage{my_commands}
\definecolor{darkbrown}{rgb}{0.4, 0.26, 0.13}

\newcommand{\valnov}{\textsc{ValNov}}

\title{Similarity-weighted Construction of Contextualized Commonsense Knowledge Graphs for Knowledge-intense Argumentation Tasks}

\author{
\\
\textbf{Moritz Plenz}$^\dag$\ \ ~~\textbf{Juri Opitz}$^\dag$\ \ ~~ \textbf{Philipp Heinisch}$^\ddag$\ \ ~~\textbf{Philipp Cimiano}$^\ddag$\ \ ~~\textbf{Anette Frank}$^\dag$ \\\\
$^\dag$Heidelberg University %
\ \ ~~$^\ddag$Bielefeld University \\
\texttt{\{plenz,opitz,frank\}@cl.uni-heidelberg.de}\hspace{0.5cm} \\
\hspace{0.5cm}\texttt{\{pheinisch,cimiano\}@techfak.uni-bielefeld.de}
\\
}

\begin{document}
\maketitle
\begin{abstract}
Arguments often do not make explicit how a conclusion follows from its premises. To compensate for this lack, we enrich arguments with structured background knowledge to support knowledge-intense argumentation tasks. We present a new \textit{un\-su\-per\-vised} method for con\-struc\-ting \textit{Contextualized Commonsense Knowledge Graphs} (CCKGs) that selects \textit{contextually relevant} knowledge from large knowledge graphs (KGs) efficiently and at high quality. Our work goes beyond con\-text-\-in\-sen\-sitive knowledge extraction heuristics by computing semantic similarity between KG triplets and textual arguments. Using these triplet similarities as weights, we extract \textit{contextualized knowledge paths} that connect a conclusion to its premise, while maximizing similarity to the argument. We combine multiple paths into a CCKG that we optionally prune to reduce noise and raise precision. Intrinsic evaluation of the quality of our graphs shows that our method is effective for (re)constructing human explanation graphs. Manual evaluations in a large-scale knowledge selection setup confirm high recall and precision of implicit CSK in the CCKGs. Finally, we demonstrate the effec\-tive\-ness of CCKGs in a knowledge-insensitive argument quality rating task, outperforming strong baselines and rivaling a GPT-3 based system.\footnote{Our code and data are available at \\ \url{https://github.com/Heidelberg-NLP/CCKG}}
\end{abstract}

\section{Introduction}
\begin{figure}
    \centering
    \includegraphics[width=\linewidth]{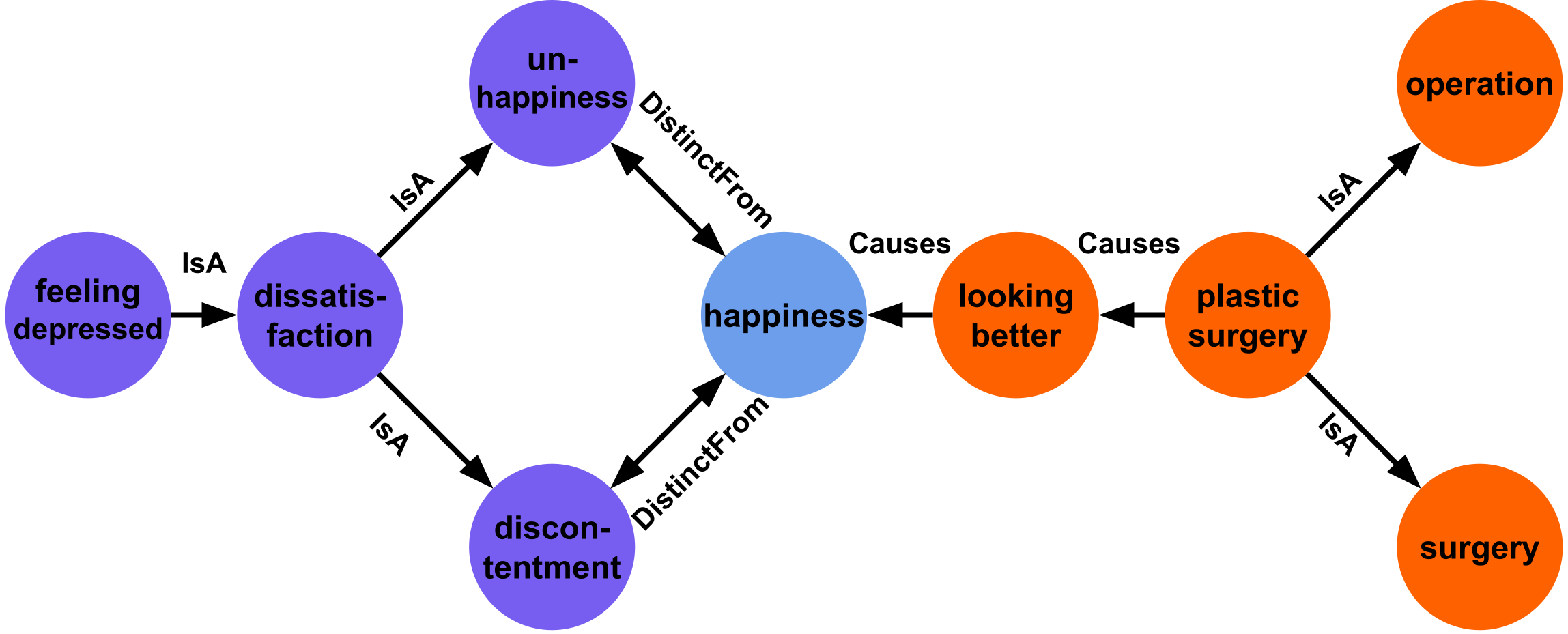}
    \caption{CCKG connecting the premise "\textit{A person is un\-happy if she is dissatisfied with her body.}" to the con\-clusion "\textit{Plastic surgery raises patients' self esteem and allows them to lead normal happy lives.}" Concepts as\-so\-cia\-ted with premise and conclusion are colored in violet and orange, respectively. The graph makes explicit that \textit{plastic surgery causes looking better}, which in turn \textit{causes happiness}, which is \textit{distinct from dissatisfaction}.}
    \label{fig:example_CCKG}
\end{figure}

Computational argumentation is a growing field with relevant applications, such as argument retrieval \citep{wachsmuth-etal-2017-building, bondarenko-etal-2021-overview}, argument analysis \citep{feng-hirst-2011-classifying,reed2014ova+,wachsmuth-etal-2017-argumentation,jo-etal-2020-detecting,opitz-etal-2021-explainable} or generation \citep{schiller-etal-2021-aspect, alshomary-etal-2021-belief,heinisch-etal-2022-strategies}. Argumentation requires deep understanding of argumentative statements and how they relate to each other. Often, 
commonsense knowledge (CSK) is needed to understand how a premise connects to its conclusion, as %
these connections are often left implicit, as shown in \F\ref{fig:example_CCKG}. 
While humans can easily infer implied knowledge, for machines extra mechanisms are needed to inject missing knowledge %
to better solve argumentative tasks \citep{moens2018argumentation, beckeretal:2020b, lauscher2021scientia,singh-etal-2022-irac}.

Methods to inject such knowledge %
either rely on 
\textit{parametric memory}, where CSK is stored in the parameters of large %
language models (LLMs), or \textit{non-parametric memory}, where CSK is stored in external knowledge bases. In the LLM approach, latent CSK can be directly exploited in downstream tasks \citep{petroni-etal-2019-language, li-etal-2021-implicit} or the LLM is fine-tuned 
to generate the CSK in explicit form \citep{bosselut-etal-2019-comet,bansal-etal-2022-cose}. However, approaches based on parametric memory have drawbacks: they often are difficult to adapt to new domains \citep{liu-etal-2022-challenges} or suffer from risk of hallucinations and unsafe generations \citep{levy-etal-2022-safetext} since they %
are not traceably grounded.

Explicit and structured CSK
is available in commonsense knowledge graphs (KGs) \citep{vrandecic-etal-2014-wikidata, speer-etal-2017-conceptnet,hwang-etal-2021-cometatomic}.
But KGs are large and \textit{not} contextualized, which makes selecting relevant knowledge difficult.

We can extract knowledge in the form of individual triplets \citep{liu-etal-2022-relational}, but this does \textit{not} allow for multi-hop reasoning over (potentially disconnected) triplets. Extracting \textit{paths} consisting of multiple triplets allows multi-hop reasoning \citep{paul-etal-2020-argumentative}, but systems cannot exploit potential interactions between multiple paths. Our approach extends the idea of multi-hop path extraction by combining multiple such paths into a graph -- our \textit{Contextualized Commonsense Knowledge Graph}. %
The CCKGs are small and tailored to a specific argument, as shown in \F\ref{fig:example_CCKG}, which makes them applicable in joint reasoning models \citep{yasunaga-etal-2022-dragon}. Similar to \textit{retrieval models} \citep{feldman-el-yaniv-2019-multi} that extract relevant passages from text for knowledge extension, our approach extracts relevant subgraphs from structured KGs.  

We can find 
connecting paths in large KGs by extracting \textit{shortest paths} that link pairs of 
concepts. But the paths are not guaranteed to provide \textit{relevant} knowledge for a given context, as intermediate triplets might be off-topic. To mitigate this problem, we compute \textit{edge weights} to rate the semantic similarity of individual KG triplets to the argument at hand, and extract \textit{weighted shortest paths} that are maximally similar to the argument. Combining the paths into a CCKG encapsulates 
relevant CSK. We compute the edge weights using SBERT 
without extra fine-tuning, and rely on graph algorithms for CCKG construction. Hence, our method is \textit{unsupervised} and applicable in zero-shot settings. 

Our main contributions are:

\textbf{i.)} We present an unsupervised \textit{Contextualized Commonsense Knowledge Graph} (CCKG) con\-stru\-ction method that enriches arguments with \textit{relevant} %
CSK, by combining 
similarity-based contextualization with graph algorithms for subgraph extraction.

\textbf{ii.)} We evaluate the \textit{quality} of CCKGs 
against manually created CSK graphs from an existing argumentation explainability task, where our method outperforms strong supervised baselines. Manual annotation shows that our CCKGs achieve high recall and precision for capturing implicit CSK. 

\textbf{iii.)} %
We evaluate our CCKGs extrinsically in a knowledge-intense argumentative transfer task. We construct CCKGs to predict the \textit{validity} and \textit{novelty} of argument conclusions, using a lightweight classifier %
which combines graph and textual 
features. We achieve strong results, rivaling a SOTA GPT-3 system and outperforming %
other supervised systems, which -- along with ablations -- demonstrates the quality, effectiveness and transparency of
CCKGs. %

\section{Background and Related Work}
When humans debate a topic, they typically leverage a vast body of \textit{background knowledge}, some already known to them and other knowledge subject to addition, e.g., by looking up a Wikipedia entry.
Therefore, with the availability of large-scale KGs \citep{auer-etal-2007-dbpedia, speer-etal-2017-conceptnet}, and with the advent of LLMs that have been shown to learn knowledge during self-supervised training \citep{bosselut-etal-2019-comet}, we observe growing interest in incorporating knowledge into computational argumentation systems \citep{beckeretal:2020b,lauscher2021scientia,singh-etal-2022-irac}. 

Of particular interest %
is the \mbox{(re-)c}onstruction of implicit \textit{commonsense knowledge} (CSK) \citep{moens2018argumentation, lawrence-reed-2020-argument, becker-etal-2021-reconstructing} within or between arguments. Usually, the goal is to improve downstream-task performance of  systems, e.g., improving argumentative relation classification by connecting concepts with paths found in KGs \citep{paul-etal-2020-argumentative}, or improving argument quality prediction by extracting KG distance features \citep{yazdi-etal-2022-kevin}. But the aim can also extend to \textit{argumentative explanations}, propelled by an emergent need for more transparency of model predictions \citep{niven-kao-2019-probing}, which is crucial for argumentative decision making \citep{ijcai2021p600}. Therefore, \citet{saha-etal-2021-explagraphs, saha-etal-2022-explanation} manually created small CSK explanation graphs and developed fine-tuned language models to generate such graphs automatically.

\begin{figure*}
    \centering
    \includegraphics[width=0.95\linewidth]{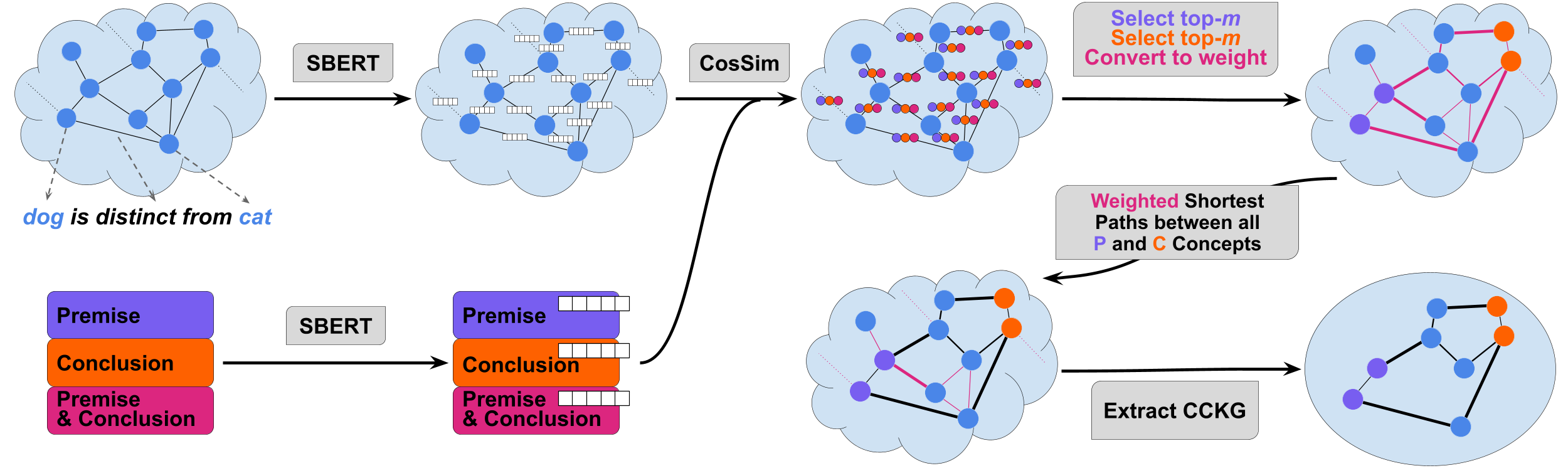}
    \caption{Overview of our method for Contextualized Commonsense Knowledge Graph (CCKG) construction.}
    \label{fig:method_overview}
\end{figure*}

Prior 
approaches for retrieving CSK 
suffer from several \textit{issues}, e.g., 
\citet{botschen-etal-2018-frame}
enrich single tokens but can't provide longer reasoning paths. 
By contrast, works
that 
construct
reasoning paths either do not exploit their interactions, are intransparent 
on which paths are used for prediction
\citep{paul-etal-2020-argumentative}, employ thresholds that are hard to tailor to different tasks \citep{9667720-li-topic-kgconstruct}, or depend on knowledge generated from LLMs \citep{becker-etal-2021-reconstructing, bansal-etal-2022-cose, saha-etal-2022-explanation}, which may decrease trust in the provided knowledge due to 
hallucinations \citep{xiao-wang-2021-hallucination, hoover-etal-2021-linguistic, 10.1145/3571730-hallucinate-survey}. In our work, we aim to unify
the strengths of such approaches while mitigating their weaknesses: Our CCKG construction
method is i) context-sensitive without requiring threshold selection, and extracts CSK graphs that 
provide ii) accurate multi-hop reasoning structures and
iii) are transparently
grounded in a large KG, and hence, iv) yield strong task performance in a transfer setting.

\section{CCKG Construction} \label{sec:method}
Given an argument, we aim to enrich it with CSK that connects the argument's premise and con\-clu\-sion via a \textit{Contextualized Commonsense Knowledge Graph} (CCKG).
\F\ref{fig:method_overview} shows an overview of our method. In a pre-processing step we compute a semantic embedding for each triplet of the KG. Using these embeddings we compute the semantic similarity of each triplet to the \textit{premise}, to the \textit{conclusion} and to the \textit{argument} as a whole. By selecting the triplets with maximal similarity scores, we obtain relevant concepts associated with the premise and conclusion. 
Next we aim to connect these concepts with relevant reasoning paths, i.e., short paths containing triplets that match the argument. We thus convert the \textit{argument} similarity scores to edge weights, and connect the extracted concepts with \textit{weighted shortest paths} that maximize the semantic similarity between the path and the argument. 
Optionally, we further prune the CCKG.\footnote{The pruning is not shown in \F \ref{fig:method_overview}.} Below we describe each step in detail. 

\paragraph{Pre-processing} 
We compute a sentence embedding for each triplet in the KG by first \textit{verbalizing} the triplets using a designated template for each relation (see \S \ref{sec:app:method_verbalization}). %
We then use a \mbox{\textit{S(entence)BERT}} \citep{reimers-gurevych-2019-sentence} encoder to compute an embedding for each verbalized triplet. Verbalization and triplet encoding is independent from the argument, so this step is
executed only once.

\paragraph{Similarity Computation} Given an \textit{argument} $A$ consisting of \textit{premise} $P$ and \textit{conclusion} $C$, we embed $P$, $C$ and $A$ = ($P$ concatenated with $C$) using SBERT. For each embedding we compute its \textit{cosine similarity} to all KG triplet embeddings. This gives us three similarity scores for each triplet: $s_P$, $s_C$ and $s_A$. %
Note that all triplet similarity scores can be computed in one single matrix multiplication, which is cheap despite a usually large number of triplets in a KG.

\paragraph{Triplet Selection for Concept Extraction} 
We select all concepts from the $m$ triplets that achieve highest similarity to $P$ and $C$, respectively, where $m$ is a hyperparameter.\footnote{This means that we extract up to $4m$ concepts from up to $2m$ triplets.} %
By using the semantic similarity of KG triplets to the textual argument as criterion for concept selection, we aim to extract concepts of higher \textit{relevance} to the argument, compared to selection by string overlap. String overlap can only capture concepts that are explicitly mentioned, which can result in incomplete extractions in case only near-synonyms appear in the KG. 
Preliminary experiments (see \S \ref{sec:app:prelim_exp_concept_extraction}) showed that computing similarity between individual concepts and the text results in concepts that are overly specific and not well connected in the KG. 
With limited connections, the shortest path search is restricted to very few paths, which can result in \textit{non-} or \textit{weakly-contextualized} paths.
Thus, we extract $P$- and $C$-concepts from selected triplets, which yields more general and better connected concepts.

\paragraph{Similarity-weighted Shortest Paths}
We use Dijkstra's algorithm \citep{dijkstra-1959-note} to find \textit{weighted shortest paths} between all pairs of extracted concepts. The algorithm requires non-negative edge weights that represent the semantic \textit{dis\-similarity} between a triplet and the argument. We therefore convert the argument similarity $s_A$ of each triplet to an edge weight $w = \nicefrac{\left( 1-s_A \right)}{2}$. %
The weighted shortest paths minimize the sum of edge weights and thus, maximize semantic similarity between the argument and the path, i.e., its constituting triplets. 

\paragraph{CCKG} We combine all weighted shortest paths to yield the final CCKG. By construction it includes 
i) $P$-concepts and $C$-concepts representing the premise and conclusion, respectively, and ii) reasoning paths that provide CSK connections between $P$ and $C$. Overall, the CCKG enriches the argument with CSK that makes the connections between premise and conclusion explicit. 

\paragraph{Pruning} 
By merging all weighted shortest paths, we obtain a graph with high coverage of potential rea\-so\-ning paths, but it may also include noise. To eliminate noise and increase precision, we optionally prune CCKGs: We rank all concepts by their semantic similarity to the argument. %
Starting with the most dissimilar concept, we greedily delete concepts from the CCKG unless the concept is i) a $P$- or $C$-concept or ii) a separator, i.e., a concept that makes the CCKG disconnected if removed. These constraints ensure that the pruned CCKG still covers both premise and conclusion, and preserves their connection. %
\F \ref{fig:example_CCKG} shows a pruned CCKG. 

\section{Experiments} \label{sec:experiments}
We evaluate our CCKG construction method \textit{intrinsically} (\S \ref{sec:intrinsic}) in an argumentative commonsense graph generation task and \textit{extrinsically} (\S \ref{sec:extrinsic}) in a knowledge-intense conclusion classification task. %

\paragraph{Experimental setup}
We instantiate our KG using the English part of ConceptNet (CN) Version 5.7 \citep{speer-etal-2017-conceptnet}, with or without the $\mathrm{RelatedTo}$ relation (see \S \ref{sec:app:knowledge_graph_statistics} for details). CN is directed, but for the shortest path search we consider all edges to be undirected. 
For SBERT similarity computation we verbalize the CN relations using templates shown in \S\ref{sec:app:method_verbalization}.
We use the Huggingface implementation\footnote{\texttt{sentence-transformers/all-mpnet-base-v2}} of SBERT. 
For shortest path search we use Dijkstra's algorithm implemented in iGraph \citep{dijkstra-1959-note,csardi-nepusz-2006-igraph}. 

\paragraph{Baselines}
Besides task-specific baselines we com\-pare to two versions of our method that ablate the edge weights, i.e., the shortest path search is unweighted and hence \textit{not} contextualized. 
We either i) randomly select one shortest path between each pair of concepts (\textbf{w/o EW\textsubscript{O}}), or ii) consider all shortest paths between two concepts (\textbf{w/o EW\textsubscript{A}}). The latter can result in large graphs\footnote{e.g., there are $\sim$100 shortest paths linking $\mathrm{person}$ to $\mathrm{work}$ in ConceptNet.} which increases computational costs in downstream tasks.

\subsection{Intrinsic evaluation on ExplaGraphs} \label{sec:intrinsic}
Our aim is to directly assess if the constructed CCKGs capture implicit CSK in arguments. This assessment is challenging, as gold data on implicit connections in argumentation is scarce and, as in any generation task, there is not only one single correct answer. To the best of our knowledge, only \citet{saha-etal-2021-explagraphs} provide relevant data. They introduce \textit{ExplaGraphs}, a \textit{generative structured commonsense reasoning task} with a corresponding dataset. Given a \textit{belief} and a \textit{support} or \textit{counter} \textit{argument}%
, the task is to generate a \textit{commonsense explanation graph} that is expected to explain the argument's \textit{stance}. %

We adapt their data to our setting of \textit{premise-conclusion} pairs by considering the \textit{argument} as a \textit{premise} and the \textit{belief} as \textit{conclusion}, which yields plausible premise-conclusion pairs for \textit{supports} (see \S \ref{sec:app:eg_manual_detailed_results}). %
For example, the 
premise and conclusion in \F \ref{fig:example_CCKG} have been constructed this way. 
Similarly, we can turn their \textit{counters} into premise-conclusion pairs. %
In this case, the belief does not form a plausible conclusion, but we can %
make their implicit knowledge connections explicit via the 
CCKG anyway. %

\citet{saha-etal-2021-explagraphs}'s gold graphs are manually constructed. Edge relations were chosen from the set of CN relations, with additional \textit{negated} relations, such as \textit{NotCapableOf}. %
Several constraints were enforced on the graphs to ensure better quality during data collection and to simplify evaluation. The graphs are connected directed acyclic graphs consisting of 3-8 triplets. Concepts have a maximum of three words and at least two concepts must stem from the premise and conclusion, respectively.

Our method does not necessarily fulfill these constraints by construction, and also cannot learn them, as it is unsupervised. Also, the imposed constraints are not necessarily beneficial for downstream tasks. We discuss appropriate metrics to compare our CCKGs against ExplaGraphs in \S \ref{sec:intrinsic_goldgraphs}. 

\begin{table*}
    \centering
     \resizebox{1.0\linewidth}{!}{
    \begin{tabular}{ccc|cccccccccc}
    \toprule
        \multicolumn{3}{c|}{Configuration} & \#nodes & \#edges & C P $\uparrow$ & C R $\uparrow$ & C F1 $\uparrow$ & T P $\uparrow$ & T R $\uparrow$ & T F1 $\uparrow$ & GED $\downarrow$ & G-BS $\uparrow$ \\
        \midrule
        \parbox[t]{2mm}{\multirow{3}{*}{\rotatebox[origin=c]{90}{CCKG}}} 
        & & $m=1$ &  4.0 &  3.0 & \textbf{52.54} & 37.94 & \textbf{42.67} & \textbf{28.55} & 19.78 & \textbf{22.13} & \textbf{0.3435} & 66.88 \\
        & & $m=2$ &  6.6 &  5.8 & 36.67 & 44.36 & 38.88 & 19.42 & 25.44 & 20.97 & 0.3745 & \textbf{74.26} \\
        & & $m=3$ &  9.2 &  8.5 & 29.25 & \textbf{48.55} & 35.49 & 15.51 & \textbf{29.63} & 19.56 & 0.4313 & 64.50 \\
        \midrule
        \parbox[t]{2mm}{\multirow{4}{*}{\rotatebox[origin=c]{90}{Supervised}}}
        & & RE-SP & 5.9 & 4.9 & 42.19 & 46.17 & 42.94 & 1.17 & 1.34 & 1.20 & 0.3706 & 74.63 \\
        & & T5    & 4.5 & 3.3 & 51.87 & 44.68 & 47.25 & 4.10 & 3.59 & 3.77 & 0.3320 & 76.26 \\
        & & max-marg. & 4.7 & 3.5 & 50.47 & 44.48 & 46.52 & 4.02 & 3.68 & 3.79 & 0.3315 & \textbf{77.96} \\
        & & contrastive & 4.6 & 3.4 & \textbf{53.70} & \textbf{46.93} & \textbf{49.26} & \textbf{5.18} & \textbf{4.75} & \textbf{4.88} & \textbf{0.3314} & 77.04 \\
        \bottomrule
    \end{tabular}}
    \caption{Intrinsic evaluation of pruned CCKGs from ExplaKnow on the ExplaGraphs dev split. 
    \textit{C P}, \textit{C R} and \textit{C F1} are precision, recall and F1 scores of concepts. \textit{T P}, \textit{T R} and \textit{T F1} are the same for triplets. \textit{GED} is normalized Graph-Edit-Distance; \textit{G-BS} is Graph-BERT-Score (see \S \ref{sec:app:saha_metrics}). All values are macro-averages across all 398 graphs.
    }
    \label{tab:intrinsic}
\end{table*}

\citet{saha-etal-2021-explagraphs}'s data consists of 2368/ 398/ 400 text-graph pairs in the train, dev and test set. 
Since the test set is not public, we report results on the dev set. We do \textit{not} need any data for hyperparameter tuning, as our method is unsupervised. 

\subsubsection{Evaluation against gold graphs} \label{sec:intrinsic_goldgraphs}
\paragraph{ExplaKnow}
Automatically assessing the semantic similarity of two graphs is challenging. Concepts in CN and ExplaGraphs are both in free-form, hence we find only few matching concepts in the two resources.
To circumvent this problem for our intrinsic evaluation, we replace CN as our external KG with an artificially created \textit{ExplaKnow} KG, which we form by combining all gold graphs from samples in the train and dev set into a single graph. %
The resulting KG has $\sim 1\%$ of CN's size, but with comparable density. Despite its smaller size, retrieving information from \textit{ExplaKnow} is non-trivial as it includes many topics, with different perspectives and stances for each of them. We hence use \textit{ExplaKnow} as a proxy to perform intrinsic quality estimation of our graph construction method against \citet{saha-etal-2021-explagraphs}'s gold graphs. \S \ref{sec:app:knowledge_graph_statistics} shows detailed statistics of \textit{ExplaKnow} vs.\ \textit{CN}. 

\paragraph{Metrics} 
We aim to assess how \textit{semantically} and \textit{structurally} similar our CCKGs are to the gold graphs, using a selection of metrics that focus on different aspects of similarity. We measure \textit{precision}, \textit{recall} and \textit{F1}-scores for \textit{concepts} and \textit{triplets}. %
\textit{Concept F1-score} indicates the ratio of correct concepts in the constructed CCKGs, as a measure of topical overlap. By contrast, the triplets encode concrete CSK statements, and hence \textit{triplet F1-score} provides a more rigid measure of the semantic overlap of a pair of graphs. 
Hence, we consider triplet F1-score as our main metric and report concept scores as additional information.
We further include two graph structure metrics from \citet{saha-etal-2021-explagraphs}: normalized graph edit distance (\textit{GED}) and G-BERTScore (\textit{G-BS}). Please refer to \S \ref{sec:app:saha_metrics} for further details on evaluation metrics applied in \citet{saha-etal-2021-explagraphs}. %

\paragraph{Baselines}
We compare against supervised methods by \citet{saha-etal-2021-explagraphs} (\textit{RE-SP}) and \citet{saha-etal-2022-explanation} (\textit{T5}, \textit{max-margin}, \textit{contrastive}). 
Their models are all trained on gold graphs in a supervised manner. 
\textbf{RE-SP} predicts concepts with fine-tuned RoBERTa and BART models and edge probabilities between concepts are predicted with another fine-tuned RoBERTa model. The system finally combines the concepts and probability-weighted edges to a graph using integer linear programming. The other baselines predict a stance with a fine-tuned RoBERTa model, then a fine-tuned T5-large model predicts the graph in a linearized form conditioned on the belief, argument and predicted stance. \textbf{T5} is fine-tuned on the training data with a standard cross-entropy loss. \textbf{Max-margin} and \textbf{contrastive} extend \textit{T5} to additionally learn from negative samples via a \textit{max-margin loss}, and from positive and negative samples via a \textit{contrastive loss}, respectively.

\subsubsection*{Automatic evaluation of CCKG on ExplaKnow}

\T\ref{tab:intrinsic} shows results for pruned CCKGs. The su\-per\-vised methods outperform CCKG by a small mar\-gin in \textit{concept metrics}. By contrast, CCKG out\-performs all supervised methods by \SI{400}{\%} and more in \textit{triplet metrics}. This in\-di\-cates that the supervised models tend to generate correct concepts, but struggle to connect them in meaningful relations.
By contrast, our approach, being grounded in a KG, attracts contextually similar triplets.

The \textit{GED} and \textit{G-BS} metrics show better results for the supervised methods, differing by \SI{1.2}{\pp} and \SI{3.7}{\pp} for the best supervised systems, respectively. However, our method matches or outperforms the RE-SP model that respects structural constraints by construction. 
Note that both metrics put high emphasis on the graph structure, which the supervised models are optimized to match. Our unsupervised method, by construction, does not necessarily fulfill %
the structural constraints that are imposed on the gold graphs, 
and cannot learn them. Hence, it is expected that the supervised models fit the 
structural constraints reflected in the train data much better. We thus consider the competitive performance of our unsupervised method as a strong result, which is confirmed by the very high triplet scores. 

Increasing $m$ ($\sim$ number of extracted concepts) increases the size of the CCKGs, which increases recall but lowers precision. The F1-scores are best for $m=1$. For downstream tasks, $m$ should be chosen according to the task, and depending on whether higher recall or higher precision is desired. 

\S \ref{sec:app:eg_further_experiments} reports further experiments which show that i) CCKGs outperform \textit{uncontextualized baselines}, also when CCKGs are constructed from \textit{ConceptNet}; %
ii) they achieve similar performance for support and counter instances; iii) verbalization of triplets 
has a small impact, but more \textit{natural verbalizations} achieve better performance; iv) using more than one weighted shortest path increases recall but decreases precision; v) pruning driven by structural features achieves comparable 
quality to pruning by semantic similarity. In \S \ref{sec:extrinsic} we introduce a variation of the CCKG construction which extracts concepts from constituents of the argument. We also test this method on ExplaGraphs in \S \ref{sec:app:eg_further_experiments}.

\subsubsection{Manual evaluation of CN subgraphs} \label{sec:intrinsic_human}
\citet{saha-etal-2021-explagraphs}'s graphs with \textit{ExplaKnow} as underlying knowledge resource offer a concise evaluation target for our CCKG construction method. But ExplaKnow is small and its concepts have been tailored to the data during instance-level data creation. To obtain a quality estimate for CCKG in a more realistic setup, we additionally conduct a manual evaluation of CCKGs on the same data, but extracted from the large \textit{ConceptNet} (CN) resource.

\paragraph{CCKGs from CN} We construct CCKGs from CN, but exclude its unspecific $\mathrm{RelatedTo}$ edges. We set $m=3$ since CN concepts are less specific compared to ExplaKnow, hence we expect that larger graphs are required to cover the targeted content. To counter-balance the larger graph size we apply pruning. In this setup, we cannot use \citet{saha-etal-2021-explagraphs}'s gold graphs as evaluation targets and therefore perform manual quality assessment.

\paragraph{Annotation}
Two independent expert annotators\footnote{Students with advanced/native competence of English.} manually labeled all 199 \textit{support instances} in the ExplaGraphs dev set. First, they assess if arguments are \textit{plausible} and include an \textit{implicit CSK connection} that links the conclusion to the premise. On the 115 instances that fulfilled both criteria unanimously, we \textbf{evaluate the quality of CCKGs}. To estimate \textbf{recall}, we inquire whether the CCKG expresses the implicit CSK that links the premise and the conclusion \textit{completely}, \textit{partially} or \textit{not at all}. Such implicit CSK can be expressed, for example, by a chain of triplets as shown in \F\ref{fig:example_CCKG}. To estimate fine-grained \textbf{precision}, the annotators had to label individual triplets as either \textit{positive} (expresses implicit CSK), \textit{neutral} (does not express implicit CSK, but matches the topic), \textit{unrelated} (does not match the topic) or \textit{negative} (contradicts implicit CSK or the conclusion)\footnote{E.g., $\triple{human\_cloning}{IsA}{considered\_unethical}$ is an example of a 
negative triplet in a CCKG for an argument that \textit{supports} cloning.}. 
This allows us to assess the precision of triplets showing \textit{implicit CSK} (positive triplets) and the precision of triplets being \textit{in-topic} (positive, neutral or negative). See \S \ref{sec:app:eg_manual_detailed_description} for detailed annotation guidelines. 

\paragraph{Results} 
\S\ref{sec:app:eg_manual_detailed_results} \T\ref{tab:app:eg_manual_detailed_results} and \ref{tab:app:eg_manual_detailed_results_macro} show detailed ana\-ly\-sis of the annotation results. We report the main findings here. 
\SI{29.57}{\%} of CCKGs were un\-ani\-mou\-sly judged to show the \textit{implicit CSK connection completely}, i.e., 
the CCKG explains the argument fully. 
This result almost doubles to \SI{59.13}{\%} when considering graphs that at least one annotator labeled as complete. \SI{88.70}{\%} show the implicit CSK \textit{partially}. Thus, CCKGs have \textbf{high recall} of implicit CSK and hence can help making implicit connections explicit. 
At the level of individual triplets, our annotation reveals that CCKGs have a \textbf{high macro triplet precision}, i.e., averaged over individual graphs, of \SI{39.43}{\%} and \SI{73.87}{\%} for \textit{showing implicit CSK} when considering unanimously labeled triplets, and triplets labeled as positive by at least one annotator, respectively. Equivalent macro precision scores for \textit{in-topic triplets} are \SI{92.76}{\%} and \SI{99.20}{\%}. This shows that a substantial amount of triplets reflects implicit CSK, and that almost all triplets are from the correct topic. Triplets from wrong topics are avoided due to strong contextualization in CCKG construction and pruning. 

We also gained qualitative insights. \textbf{Missing knowledge}: We find cases of arguments on a topic that lacks coverage in CN, resulting in similar CCKGs for different arguments.\footnote{18 out of 22 instances on \textit{entrapment} yield identical CCKGs, 
due to lack of coverage in CN.}
\textbf{Ambiguity}: CN concepts are not disambiguated. A path may thus run through concepts that take different senses, making the path meaningless.\footnote{For example, the following chain of triplets $\left(\mathrm{river\_bank},\,\mathrm{IsA},\,\mathrm{bank},\,\mathrm{UsedFor},\,\mathrm{keeping\_money\_safe}\right)$, is a path that connects the concepts $\mathrm{river\_bank}$ and $\mathrm{keeping\_money\_safe}$, and is established by the intermediary concept $\mathrm{bank}$ that takes a different meaning in the two constituting triplets.}

\subsection{Extrinsic evaluation: Predicting Validity and Novelty of Arguments (\valnov)} \label{sec:extrinsic}
We now investigate the \textit{effectiveness} of CCKGs -- used to explicate implicit CSK in arguments -- in the novel, knowledge-intense argumentation task \valnov. %
We evaluate the \textit{robustness} of our unsupervised method relying on non-parametric knowledge, compared to su\-per\-vised graph generation systems applied out-of-domain, as well as SOTA \valnov~systems. 

\paragraph{Task description}
\citet{heinisch-etal-2022-overview} introduced a novel argument inference task \valnov~as a community shared task. Given a textual premise and conclusion, the task is to predict whether the conclusion is i) \textit{valid} and ii) \textit{novel} with respect to its premise. A conclusion is \textit{valid} if it is \textit{justified} by the premise. It is \textit{novel} if it contains premise-related content that is not part of the premise, i.e. the conclusion \textit{adds novel content} to the premise. Please refer to \S \ref{sec:app:vn_datastatistics} for data statistics. 

Systems are expected to report macro F1-\-scores for joint and individual prediction of validity and novelty. In joint modeling we dis\-tin\-guish 4 classes: i) \textit{valid \& novel}, ii) \textit{non-valid \& novel}, iii) \textit{valid \& non-novel}, iv) \textit{non-valid \& non-\-novel}. %
The training data is unbalanced with respect to these 4 classes. 

\paragraph{Predicting Validity and Novelty from CCKGs}
We hypothesize that CCKGs show structural characteristics that correlate with validity and novelty: For instance, a \textit{valid} conclusion should be well connected to its premise in the constructed CCKG, and a \textit{novel} conclusion should result in a CCKG with long paths from the premise to its conclusion. To test these hypotheses we extract graph features from the CCKGs and combine them with textual features from the argument. We feed all features to shallow classifiers to predict the validity and novelty of conclusions. 
Note that interaction between the CCKG and the argument is limited in this approach, which allows us to isolate and investigate the expressiveness of our CCKGs. 

\textbf{CCKG details} 
The \valnov~dataset contains arguments that are relatively long (76 tokens in avg.), often comprising more that one aspect/ perspective. %
This negatively effects the quality of triplet selection for concept extraction: the extracted concepts are semantically relevant, but often don't span the entire argument. Thus, we parse the text into constituents and select concepts from the top-$m$ triplets for each %
constituent individually. 

Pruning CCKGs completely bears the danger of removing relevant structural aspects of CCKGs. We therefore experiment with \textit{partial pruning}, that only removes the most dissimilar prunable concepts. This enables a more fine-grained balance of recall and precision compared to complete pruning.

We obtain best performance using parsing, partial pruning (\SI{75}{\%}), $m=2$ and CN w/o $\mathrm{RelatedTo}$. Please refer to \S \ref{sec:app:valnov_modelvariations} for further details on concept extraction with parsing and partial pruning. 

\textbf{Feature extraction:} We extract 15 graph features from each CCKG: 5 characterizing its \textit{size}, 6 its \textit{connectivity} and 4 the \textit{distance between premise and conclusion in the CCKG}. As textual features we use the \textit{semantic similarity} of premise and conclusion, and predictions from a \textit{NLI}-model. %
We obtain 19 features in total. See \S \ref{sec:app:valnov_featureextraction} for detailed description of the features. 

\textbf{Classifier} 
We train Random Forests and SVMs in a multi-class setting, considering validity and novelty jointly. 
Following \citet{yazdi-etal-2022-kevin} we use upsampling to balance the training data. 
Results are averaged over 5 different runs. %
Please refer to \S \ref{sec:app:valnov_classifiers} for hyperparameters and implementation details of the classifiers.  

\paragraph{Baselines}
We compare to su\-per\-vi\-sed ExplaGraphs generation systems by embedding their graphs into our classifier, and to systems participating in the \valnov~shared task: the two best-\-per\-for\-ming submissions, the System-Average (average of all submissions) and the ST baseline.

We evaluate against \textbf{supervised graph construction methods} \citep{saha-etal-2022-explanation} (see \S \ref{sec:intrinsic_goldgraphs}), to assess their performance in an out-of-domain setting, compared to our unsupervised CCKG construction method. We apply their trained graph generation models to \valnov~arguments and use the generated graphs exactly as we do for our CCKGs: we extract the same features to train the shallow classifier models, following our training protocol. 
Unlike our general-purpose CCKGs, these methods were trained to generate graphs for stance-classification tasks. Nevertheless, we can apply these methods to \valnov~as further baselines. 

The shared task winning system \textbf{ST-1st} \citep{meer-etal-2022-blend} prompted GPT-3 for \textit{validity} and separately fine-tuned a RobERTa-based NLI model, further enhanced with contrastive learning, for \textit{novelty}. The second-best shared task system \textbf{ST-2nd} \citep{yazdi-etal-2022-kevin} is a FFNN trained with upsampling that combines diverse features from NLI predictions, semantic similarity, predictions of validity and novelty and structural knowledge extracted from WikiData. 
The shared task baseline \textbf{BL} consists of two RoBERTa models, fine-tuned for validity and novelty separately. 

Our system resembles the \textit{ST-2nd} approach, however, their system strongly emphasizes textual features, even leveraging a fine-tuned BERT predicting validity and novelty based on text alone, and considers only two structural features from un\-con\-textualized WikiData paths. Our model, by contrast, relies on a minimal amount of textual features, leveraging standard pre-trained models without task-dependent fine-tuning. %
Hence, it strongly relies on graph features, building on the strong contextualization of CCKGs to the textual argument. %

\paragraph*{Results}
\T\ref{tab:valnov} shows the results
on the \valnov~test set. Our system CCKG achie\-ves the second best result in all metrics: \textit{validity}, \textit{novelty} and \textit{joint} prediction. Best scores are achieved either by ST-1st with GPT-3 on \textit{joint} and \textit{validity} prediction or by \citet{saha-etal-2022-explanation}'s T5 model for \textit{novelty}. Yet our approach outperforms these systems in the respective complementary metrics: \textit{novelty} for ST-1st and \textit{validity} for T5. %
CCKG clearly
outperforms T5 in joint F1 by \SI{6.2}{\pp}

\begin{table}
    \centering
         \resizebox{1.0\linewidth}{!}{
    \begin{tabular}{ccc|ccc}
    \toprule
        \multicolumn{3}{c|}{Systems and BLs} & joint F1 & Val F1 & Nov F1 \\
        \midrule \midrule
        \parbox[t]{2mm}{\multirow{4}{*}{\rotatebox[origin=c]{90}{ST}}} %
        & & 1st (GPT-3)& \textbf{\underline{45.16}} & \textbf{\underline{74.64}} & 61.75 \\
        & & 2nd (w/ KG)& 43.27 & 69.80 & 62.43 \\
        & & System Avg & 35.94 & 62.74 & 52.97 \\
        & & Baseline & 23.90 & 59.96 & 36.12 \\
        \midrule
        \parbox[t]{2mm}{\multirow{3}{*}{\rotatebox[origin=c]{90}{EG w/}}} & \parbox[t]{2mm}{\multirow{3}{*}{\rotatebox[origin=c]{90}{Ours}}} 
        & T5 & 37.71 & 67.07 & \textbf{\underline{63.53}} \\ 
        & & max-margin & 36.22 & 67.61 & 63.27 \\ 
        & & contrastive & 37.82 & 64.77 & 59.96 \\
        \midrule
        & & CCKG (Ours) & \textbf{43.91} & \textbf{70.69} & \textbf{63.30} \\
        \midrule \midrule
        \parbox[t]{2mm}{\multirow{7}{*}{\rotatebox[origin=c]{90}{Ablation}}} %
        & & w/o Graph feats. & -11.65 & -3.40 & -5.12 \\
        & & w/o Text feats. & -20.65 & -20.74 & -17.69 \\
        & & w/o EW\textsubscript{O} & -6.51 & -3.80 & -1.69 \\
        & & w/o EW\textsubscript{A} & -3.25 & -4.45 & 1.76 \\
        & & string matching & -6.71 & -3.23 & 0.55 \\
        & & w/o connectivity feats. & -5.60 & -4.01 & -0.60 \\
        & & w/o PC-distance feats. & -2.27 & -0.27 & -3.73 \\
        \bottomrule
    \end{tabular}}
    \caption{Results on \valnov: \textit{joint}, \textit{validity} and \textit{novelty} F1-scores. We compare against Shared Task (\textit{ST}) results and ExplaGraphs generation models, integrated in our \valnov~classifier (\textit{EG w/ Ours}). 
    Ablated scores are relative to \textit{CCKG~(Ours)}. 
    }
    \label{tab:valnov}
\end{table}

\citet{heinisch-etal-2022-overview}'s analysis of the \valnov~results concludes that i) LLMs are powerful predictors of \textit{validity}, due to the textual inference capabilities they acquire in pretraining on massive text sources. 
At the same time, ii) LLMs were shown to lag behind knowledge-based systems in \textit{novelty} prediction. \textit{Validity} was overall easier to solve than \textit{novelty}, and systems that performed well for \textit{novelty} had poor \textit{validity} results, and vice versa.\footnote{%
For example, prompting GPT-3 for novelty resulted in only \SI{46.07}{\%} F1 score. }
It is therefore  no surprise that our system cannot compete with GPT-3 for \textit{validity}. 
However, it achieves 2nd best performance on \textit{validity} at a high level of \SI{70.69}{\%} F1 without sacrificing \textit{novelty}. Leveraging structural knowledge, T5 achieves highest scores for novelty, but performs poorly in validity, and hence, only ranks 5th in the joint ranking. CCKGs perform well in both, validity and novelty, with one unified approach, unlike ST-1st. Our strong joint score of \SI{43.72}{\%} only gets surpassed by ST-1st, which leverages two independent systems for validity and novelty. Thus, simple feature extraction from CCKGs achieves interpretable and yet compatible scores. Our ablation will show that this is possible due to strong contextualization in the graph construction. 

\paragraph{Ablation} 
Removing graph or text features from CCKG (ours) reduces performance by \SI{11.65}{\pp} and \SI{20.65}{\pp}, respectively. %
The text is more important for \textit{validity}, while the graph has a larger impact on \textit{novelty}. Yet, both metrics benefit from both modalities. This indicates that text and CCKG contain complementary information and should be considered jointly in future work.

Ablating all edge weights incurs considerable performance losses for \textit{validity} and \textit{joint} F1. \textit{Novelty} is less affected, which shows that \textit{contextualization} is more relevant for validity. We can also empoverish contextualization by extracting concepts via string matching. This decreases performance by \SI{6.71}{\pp}, again with a larger decrease for validity. 

Feature ablation confirms that %
connectivity features are most relevant for validity, while %
premise-conclusion distance 
in the CCKG is most relevant for novelty. Further ablations %
are shown 
in \S \ref{sec:app:valnov_ablation}.

\section{Conclusion}
In this work we proposed an unsupervised method to construct \textit{Contextualized Commonsense Knowledge Graphs} (CCKGs). Our extensive evaluations show that CCKGs are of high quality, outperform context-insensitive baselines and rival strong supervised graph construction methods on diverse argumentation tasks, while offering increased robustness. Being grounded in a KG, the information captured in our CCKGs is traceable and hence interpretable. 
Future work could explore incorporation of more specific KGs to address particular domains. Using our compact, high-quality CCKGs in stronger interaction with LLMs is %
another step to address in future work. 

\section*{Limitations}
In principle our method is applicable in many domains, for example, one could use a biomedical knowledge graph instead of ConceptNet in a relevant domain. However, in this paper we only evaluate the quality of our approach in argumentative tasks which require commonsense knowledge. Our approach is unsupervised, but its performance depends on the quality of the used knowledge graph and SBERT model. %

Similarly, we only evaluate CCKGs for English data, although our approach is not limited to English if one uses multilingual SBERT models \citep{reimers-gurevych-2020-making} or a multilingual knowledge graph. 

Finally, our approach is purely extractive and hence, is limited by the coverage and quality of knowledge graphs. However, improving knowledge graphs is an active field of research and hence, high-quality and high-coverage knowledge graphs are to be expected. Furthermore, our extracted CCKGs could be augmented with generative models if coverage in the knowledge graph is not sufficient. However, that would reduce the interpretability that our approach provides. 

\section*{Ethical Considerations}
Our method extracts subgraphs from knowledge graphs. Hence, any potential biases present in the knowledge graph can propagate to our CCKGs. While this can be problematic, our approach allows to trace biases back to their origin. This is comparable to manual information extraction, as all knowledge sources can contain biases -- for example political tendencies in newspapers. 
Strategies to automatically avoid biases \citep{mehrabi-etal-2021-lawyers} could also be incorporated in future work. However, as our approach is a pure extraction, it can not generated new potentially harmful information. Thus, CCKGs are perhaps more reliable for sensitive application than knowledge representations generated without grounding.

\section*{Acknowledgements}
We want to thank Swarnadeep Saha for generating the supervised graphs (T5, max-margin and contrastive) which we compare to in \S \ref{sec:extrinsic}. We also thank our annotators for their support.  

This work was funded by DFG, the German Research Foundation, within the project ACCEPT, as part of the priority program "Robust Argumentation Machines" (RATIO,  SPP-1999).

\FloatBarrier
\bibliography{anthology,custom}
\bibliographystyle{acl_natbib}

\FloatBarrier
\appendix

\section{Method}
\subsection{Preliminary experiments on concept extraction} \label{sec:app:prelim_exp_concept_extraction}
As a preliminary experiment, we test how to extract concepts that are well-connected in the KG. 
Concepts which are not well-connected have limited options to be connected to each other, which hinders contextualization in the shortest path search. Hence, we require well-connected concepts which are not overly specific. We estimate the connectivity and specificity of concepts by their degree and number of words, respectively. 

We experiment with i) extracting concepts that are most similar to the text, and ii) extracting all concepts from the triplets that are most similar to the text. In each case we measure similarity between the concept / triplet and the text with the same SBERT model. As KG we use ConceptNet (CN) without $\mathrm{RelatedTo}$ triplets (please refer to \S \ref{sec:app:method_relatedto} for further context on the choice of the KG). 

\T\ref{tab:app:preliminary_conceptextraction_conceptvstriplet} shows the macro averages over the development split of ExplaGraphs \citep{saha-etal-2021-explagraphs} for $m=1$. Varying $m$ only has a small impact on the results. Extracting concepts via ranking triplets results in shorter concepts with high degrees, i.e. general and well-connected concepts. Thus, we extract concepts by first ranking triplets and then selecting all concepts in the top-$m$ triplets. 

\begin{table}
    \centering
    \begin{tabular}{c|rr}
        \toprule
        metric & concept & triplet \\
        \midrule
        number of words & 2.42 & 1.83 \\
        degree & 4.21 & 103.39 \\
        \bottomrule
    \end{tabular}
    \caption{Comparison of direct concept extraction and concept extraction via triplet ranking. Values are averages over extracted concept from dev set of ExplaGraphs for $m=1$. }
    \label{tab:app:preliminary_conceptextraction_conceptvstriplet}
\end{table}

\section{Experiments}
\subsection{Experimental setup}
\subsubsection{Discussion of RelatedTo in CN} \label{sec:app:method_relatedto}
More than half of all triplets in CN have the relation $\mathrm{RelatedTo}$ (see \T\ref{tab:app:knowledge_graph_statistics}). This is a very general relation and thus might cause a high degree of semantically vacuous connections. Hence, paths constructed from CN without $\mathrm{RelatedTo}$ are potentially longer, but more explicit and therefore also more expressive. On the other hand, $\mathrm{RelatedTo}$ might be necessary to make certain connections in CN. Thus we experiment with two different versions of CN: one with $\mathrm{RelatedTo}$ and one without $\mathrm{RelatedTo}$. 

To create a graph from CN excluding the $\mathrm{RelatedTo}$ relation, we first remove all triplets with this relation and then all concepts with degree 0. \T\ref{tab:app:knowledge_graph_statistics} shows statistics of CN with and without $\mathrm{RelatedTo}$. 

\subsubsection{Triplet verbalization} \label{sec:app:method_verbalization}
SBERT was pre-trained on natural language sentences, and thus is not ideal for capturing semantics of triplets. We could fine-tune SBERT to learn triplet-representations, but that might reduce the generalizability of our model. Therefore we prefer to convert the triplets to natural language, which can be processed by SBERT without any fine-tuning. 

To translate triplets to natural language we designed \textit{natural} templates that preserve the relation's meaning, but are more natural. To analyze the impact of the verbalization templates we also created \textit{static} templates, which are closer to the original relations. Our templates are shown in \T\ref{tab:app:verbalization_templates_cn} and \T\ref{tab:app:verbalization_templates_explaknow} for CN and ExplaKnow. 

Note that these templates can propagate grammatical errors from the triplets, e.g. $\triple{humans}{Desires}{freedom}$ would get verbalized to \textit{humans desires freedom} instead of the grammatically correct \textit{humans desire freedom}. In principle, automatically correcting these errors could be included in the pre-processing step of our method, but for simplicity we refrained from doing so. 

\begin{table}
    \centering
    \resizebox{.93\linewidth}{!}{
    \begin{tabular}{c|cc}
    \toprule
    Relation & Natural & Static \\
    \midrule
    RelatedTo & is related to & is related to\\
    IsA & is a & is a \\
    FormOf & is a form of & is a form of \\
    CapableOf & is capable of & is capable of\\
    MotivatedByGoal & is motivated by & is motivated by the goal \\
    HasContext & has context & has the context \\
    HasPrerequisite & has prerequisite & has the prerequisite \\
    Synonym & is a synonym of & is a synonym of\\
    Antonym & is an antonym of & is an antonym of \\
    AtLocation & is in & is at the location \\
    Desires & desires & desires \\ 
    UsedFor & is used for & is used for \\ 
    HasSubevent & has subevent & has the subevent \\ 
    HasProperty & is & has the property \\
    PartOf & is a part of & is a part of \\
    DefinedAs & is defined as & is defined as \\
    HasA & has & has a \\
    MannerOf & is a manner of & is a manner of\\
    Causes & causes & causes \\
    HasFirstSubevent & starts with & has the first subevent \\
    HasLastSubevent & ends with & has the last subevent \\
    ReceivesAction & $\star\,$can be done to & receives the action \\
    InstanceOf & is an instance of & is an instance of \\
    NotCapableOf & is not capable of & is not capable of \\
    CausesDesire & causes desire & causes the desire \\
    DistinctFrom & is distinct from & is distinct from \\
    NotDesires & does not desire & does not desire \\
    MadeOf & is made of & is made of \\
    Entails & entails & entails \\
    CreatedBy & is created by & is created by \\
    NotHasProperty & is not & does not have the property \\
    LocatedNear & is near & is located near \\
    SymbolOf & is a symbol of & is a symbol of \\    
    \bottomrule
    \end{tabular}}
    \caption{Verbalization templates for all relations in CN. \\
    $\star$: the order of concepts is inverted in the verbalization, i.e. $\triple{A}{ReceivesAction}{B}$ is verbalized as \textit{B can be done to A}. }
    \label{tab:app:verbalization_templates_cn}
\end{table}

\begin{table}
    \centering
    \resizebox{.93\linewidth}{!}{
    \begin{tabular}{c|cc}
    \toprule
    Relation & Natural & Static \\
    \midrule
    IsA & is a & is a \\
    IsNotA & is not a & is not a \\
    CapableOf & is capable of & is capable of \\
    NotCapableOf & is not capable of & is not capable of \\
    HasContext & has context & has context \\
    NotHasContext & does not have context & does not have context \\
    SynonymOf & is a synonym of & is a synonym of \\
    AntonymOf & is an antonym of & is an antonym of \\
    AtLocation & is in & is at the location \\
    NotAtLocation & is not in & is not at the location \\
    Desires & desires & desires \\
    NotDesires & does not desire & does not desire \\
    UsedFor & is used for & is used for \\
    NotUsedFor & is not used for & is not used for \\
    HasSubevent & has subevent & has subevent \\
    NotHasSubevent & does not have subevent & does not have subevent \\
    HasProperty & is & has the property \\
    NotHasProperty & is not & does not have the property \\
    PartOf & is a part of & is a part of \\
    NotPartOf & is not a part of & is not a part of \\
    Causes & causes & causes \\
    NotCauses & does not cause & does not cause \\
    ReceivesAction & $\star\,$can be done to & receives the action \\
    NotReceivesAction & $\star\,$can not be done to & does not receive the action \\
    MadeOf & is made of & is made of \\
    NotMadeOf & is not made of & is not made of \\
    CreatedBy & is created by & is created by \\
    NotCreatedBy & is not created by & is not created by \\
    \bottomrule
    \end{tabular}}
    \caption{Verbalization templates for all relations in ExplaKnow. \\
    $\star$: the order of concepts is inverted in the verbalization, e.g. $\triple{A}{NotReceivesAction}{B}$ is verbalized as \textit{B can not be done to A}. }
    \label{tab:app:verbalization_templates_explaknow}
\end{table}

\subsection{ExplaGraphs automatic evaluation}

\subsubsection{Knowledge Graph statistics} \label{sec:app:knowledge_graph_statistics}
For the statistics in \T\ref{tab:app:knowledge_graph_statistics} we consider the KGs as multi-graphs, i.e. two triplets which differ only by their relation are considered as two separate edges. The table shows statistics for ConceptNet with and without the $\mathrm{RelatedTo}$ Relation (see \S \ref{sec:app:method_relatedto}) and for ExplaKnow, the artificial KG constructed from ExplaGraphs. The average number of words is the average across all concepts in the graph. 

The table shows that ExplaKnow is smaller than CN, but has a comparable average degree. However, concepts in ExplaKnow have more words than CN's concepts on average. The intersection scores show that only \SI{35}{\%} of concepts in ExplaKnow are contained in CN, and less than \SI{1}{\%} of ExplaKnow triplets are in CN. 

\begin{table*}
    \centering
    \begin{tabular}{c|rrcccc}
        \toprule
        Knowledge Graph & \# concepts & \# triplets & avg. degree & avg. \# words & $\cap$ concepts & $\cap$ triplets \\
        \midrule
        ExplaKnow & 7,267 
        &   11,437 & 3.1 & 2.1 & 0.35 & 0.00 \\
        CN w/o RelatedTo & 939,836 & 1,313,890 & 2.8 & 1.6 & 1.00 & 1.00 \\
        CN w/ RelatedTo & 1,134,506 & 3,017,472 & 5.3 & 1.6 & 1.00 & 1.00 \\
        \bottomrule
    \end{tabular}
    \caption{Knowledge graph statistics. \textit{avg. \# words} is the average number of words per concept; \textit{$\cap$ concepts} and \textit{$\cap$ triplets} are the number of concepts and triplets respectively in the intersection between the KG and \textit{CN w/ RelatedTo} normalized by the number of concepts / triplets in the respective KG. }
    \label{tab:app:knowledge_graph_statistics}
\end{table*}

\subsubsection{Metrics proposed in \citet{saha-etal-2021-explagraphs}} \label{sec:app:saha_metrics}
\citet{saha-etal-2021-explagraphs} propose evaluation of constructed graphs in three steps, where the first two steps evaluate if the \textit{stance} is correctly predicted, and if the graph is \textit{structurally} correct, i.e. if it fulfills the structural constraints imposed by \citet{saha-etal-2021-explagraphs}. Graphs are only evaluated in the third step, if the stance-prediction and the structure are correct. In this work, we do not focus on stance prediction and also do not aim at fulfilling the artificial structural constraints. Hence, we skip the first two stages and evaluate our metrics on all graphs, independent of the predicted stance and structural constraints. 

In their third evaluation stage, \citet{saha-etal-2021-explagraphs} consider four metrics. However, two of them are automatically assessed by fine-tuned LLMs. These LLMs were fine-tuned on graphs which fulfill the structural constraints, and hence, we would have to use the LLMs out-of-domain if we were to apply them to our CCKGs. Thus, we can not rely on these automatic metrics for our graphs. However, we do adopt the other two proposed metrics from stage three:  Graph edit distance (\textbf{GED}) measures the minimal number of \textit{edits} to make two graphs isomorph. Edits are local changes, i.e. relabeling, adding or removing a concept or an edge. For increased consistency the GED is normalized to range from 0 to 1. G-BERTScore (\textbf{G-BS}) is an extension of BERTScore \citep{zhang-etal-2020-bertscore} to graphs. Triplets are considered as sentences, and BERTScore is used as an alignment score between each pair of triplets. G-BS is computed from the best alignment between the two graphs given the alignment scores.

\subsubsection{Additional experiments} \label{sec:app:eg_further_experiments}
This section shows experiments that are slight variations to the setting presented in \T\ref{tab:intrinsic}. Hence, unless stated otherwise, all CCKGs are pruned CCKGs constructed from ExplaKnow. 

\paragraph{Uncontextualized CCKG baselines}
\T\ref{tab:app:eg_baselines} shows the CCKGs and pruned CCKGs compared to the uncontextualized baselines. The results show that CCKGs outperform the baselines without edge weights in concept and triplet precision and F1, as well as in GED and G-BS. Pruning by SBERT similarity introduces contextualization to the baselines, which allows w/o EW\textsubscript{O} (i.e. only one randomly chosen unweighted shortest path between two concepts) to achieve comparable performances to the pruned CCKGs. In triplet F1 score the pruned baseline achieves the best result, but it is only outperforming the pruned CCKGs by insignificant \SI{0.09}{\pp} 

The baselines achieve increased recall compared to CCKGs, but the baselines also produce larger graphs which explains the improvements. 

\begin{table*}
\centering
     \resizebox{.9\linewidth}{!}{
    \begin{tabular}{ccc|rrrrrrrrrr}
    \toprule
        \multicolumn{3}{c|}{Configuration} & \#nodes & \#edges & C P $\uparrow$ & C R $\uparrow$ & C F1 $\uparrow$ & T P $\uparrow$ & T R $\uparrow$ & T F1 $\uparrow$ & GED $\downarrow$ & G-BS $\uparrow$ \\
        
        \midrule \midrule
        
        \parbox[t]{2mm}{\multirow{6}{*}{\rotatebox[origin=c]{90}{CCKG}}} 
        & & $m=1$ &  4.1 &  3.2 & \textbf{52.10} & 38.28 & \textbf{42.58} & \textbf{28.12} & 20.19 & \textbf{22.02} & \textbf{0.3458} & 66.41 \\
        & & $m=2$ &  7.1 &  6.6 & 35.90 & 45.40 & 38.53 & 18.68 & 26.60 & 20.55 & 0.3872 & \textbf{71.39} \\
        & & $m=3$ & 10.1 & 10.3 & 28.26 & \textbf{49.96} & 34.81 & 14.43 & \textbf{31.11} & 18.61 & 0.4524 & 60.53 \\
        \cline{2-13}
        & \parbox[t]{2mm}{\multirow{3}{*}{\rotatebox[origin=c]{90}{pruned}}}
          & $m=1$ & 4.0 & 3.0 & \textbf{52.54} & 37.94 & \textbf{42.67} & \textbf{28.55} & 19.78 & \textbf{22.13} & \textbf{0.3435} & 66.88 \\
        & & $m=2$ & 6.6 & 5.8 & 36.67 & 44.36 & 38.88 & 19.42 & 25.44 & 20.97 & 0.3745 & \textbf{74.26} \\
        & & $m=3$ & 9.2 & 8.5 & 29.25 & \textbf{48.55} & 35.49 & 15.51 & \textbf{29.63} & 19.56 & 0.4313 & 64.50 \\
        
        \midrule \midrule
        
        \parbox[t]{2mm}{\multirow{6}{*}{\rotatebox[origin=c]{90}{w/o EW\textsubscript{A}}}} 
        & & $m=1$ &  5.5 &  6.1 & \textbf{47.97} & 40.22 & \textbf{40.91} & \textbf{24.95} & 22.85 & \textbf{20.88} & \textbf{0.3805} & \textbf{61.47} \\
        & & $m=2$ & 11.4 & 16.2 & 29.76 & 49.34 & 34.27 & 14.02 & 31.96 & 16.94 & 0.4811 & 54.21 \\
        & & $m=3$ & 18.2 & 28.5 & 21.49 & \textbf{55.49} & 28.76 &  9.70 & \textbf{39.11} & 13.88 & 0.5829 & 38.96 \\
        \cline{2-13}
        & \parbox[t]{2mm}{\multirow{3}{*}{\rotatebox[origin=c]{90}{pruned}}}
          & $m=1$ & 4.0 & 3.1 & \textbf{52.36} & 37.63 & \textbf{42.46} & \textbf{28.16} & 19.80 & \textbf{22.01} & \textbf{0.3455} & 67.61 \\
        & & $m=2$ & 6.7 & 6.2 & 36.48 & 44.33 & 38.72 & 18.73 & 26.06 & 20.64 & 0.3799 & \textbf{72.54} \\
        & & $m=3$ & 9.3 & 9.5 & 28.95 & \textbf{48.69} & 35.24 & 14.74 & \textbf{30.71} & 19.05 & 0.4406 & 61.00 \\
        
        \midrule \midrule
        
        \parbox[t]{2mm}{\multirow{6}{*}{\rotatebox[origin=c]{90}{w/o EW\textsubscript{O}}}}
        & & $m=1$ &  4.6 &  4.3 & \textbf{49.77} & 39.21 & \textbf{41.77} & \textbf{26.14} & 21.91 & \textbf{21.56} & \textbf{0.3600} & \textbf{64.98} \\
        & & $m=2$ &  8.9 & 10.7 & 32.48 & 47.81 & 36.45 & 15.50 & 29.90 & 18.31 & 0.4393 & 60.42 \\
        & & $m=3$ & 13.7 & 18.3 & 24.35 & \textbf{53.71} & 31.75 & 11.30 & \textbf{36.70} & 15.89 & 0.5286 & 45.63 \\
        \cline{2-13}
        & \parbox[t]{2mm}{\multirow{3}{*}{\rotatebox[origin=c]{90}{pruned}}}
          & $m=1$ & 3.9 & 3.1 & \textbf{52.51} & 37.78 & \textbf{42.61} & \textbf{28.38} & 19.98 & \textbf{22.22} & \textbf{0.3441} & 67.75 \\
        & & $m=2$ & 6.6 & 6.2 & 36.70 & 44.55 & 38.95 & 18.92 & 26.19 & 20.83 & 0.3786 & \textbf{72.71} \\
        & & $m=3$ & 9.2 & 9.3 & 29.05 & \textbf{48.60} & 35.33 & 14.80 & \textbf{30.61} & 19.11 & 0.4390 & 61.28 \\
    \bottomrule
    \end{tabular}
    }
    \caption{Intrinsic evaluation of pruned CCKGs constructed from ExplaKnow. \textit{w/o EW\textsubscript{A}} and \textit{w/o EW\textsubscript{O}} are the baselines with unweighted shortest paths described in \S \ref{sec:experiments}. 
    }
    \label{tab:app:eg_baselines}
\end{table*}

\paragraph{CN as KG}
\T\ref{tab:app:eg_cn} shows the results when using ConceptNet (CN) as KG instead of ExplaKnow. Scores have an upper bound due to the small overlap between CN and the gold graphs (see \S \ref{sec:app:knowledge_graph_statistics}). Especially for triplets only very low scores are possible. 

However, the results show that CCKGs outperform the baselines without edge weights in concept and triplet precision and F1, as well as in GED and G-BS. The performance gap is especially prominent when comparing the unpruned versions. This is likely because the pruning by SBERT similarity introduces contextualization into the otherwise uncontextualized baselines. 

The w/o EW\textsuperscript{A} baselines (i.e. all unweighted shortest paths between two concepts) outperforms CCKGs in terms of recall, but the baseline graphs are also many times larger which greatly harms the precision. 

This confirms that CCKGs perform well in the intrinsic evaluation, also when they are constructed from CN. 

\begin{table*}
\centering
     \resizebox{.9\linewidth}{!}{
    \begin{tabular}{cccc|rrrrrrrrrr}
    \toprule
        \multicolumn{4}{c|}{Configuration} & \#nodes & \#edges & C P $\uparrow$ & C R $\uparrow$ & C F1 $\uparrow$ & T P $\uparrow$ & T R $\uparrow$ & T F1 $\uparrow$ & GED $\downarrow$ & G-BS $\uparrow$ \\
        \midrule \midrule
        \parbox[t]{2mm}{\multirow{12}{*}{\rotatebox[origin=c]{90}{CCKG}}} & \parbox[t]{2mm}{\multirow{6}{*}{\rotatebox[origin=c]{90}{w/o RT}}} 
        & & $m=1$ &  4.4 &   3.4 & \textbf{20.03} & 14.03 & \textbf{15.40} & \textbf{0.30} & 0.22 & 0.24 & \textbf{0.4393} & 57.59 \\
        & & & $m=2$ &  8.5 &   8.3 & 12.91 & 17.13 & 13.63 & 0.24 & \textbf{0.38} & \textbf{0.27} & 0.4980 & \textbf{59.51} \\
        & & & $m=3$ & 12.9 &  14.0 &  9.79 & \textbf{19.52} & 12.11 & 0.19 & \textbf{0.38} & 0.23 & 0.5762 & 49.75 \\
        \cline{3-14}
        & & \parbox[t]{2mm}{\multirow{3}{*}{\rotatebox[origin=c]{90}{pruned}}}
            & $m=1$ &  4.2 &   3.0 & \textbf{20.54} & 14.03 & \textbf{15.73} & \textbf{0.30} & 0.22 & 0.24 & \textbf{0.4314} & 59.27 \\
        & & & $m=2$ &  7.5 &   6.6 & 13.75 & 16.95 & 14.32 & 0.25 & \textbf{0.32} & \textbf{0.27} & 0.4737 & \textbf{64.35} \\
        & & & $m=3$ & 10.8 &  10.4 & 10.80 & \textbf{19.04} & 13.05 & 0.19 & \textbf{0.32} & 0.22 & 0.5389 & 56.35 \\
        \cline{2-14}
         & \parbox[t]{2mm}{\multirow{6}{*}{\rotatebox[origin=c]{90}{w/ RT}}}
          & & $m=1$ &  4.2 &   3.4 & \textbf{22.27} & 15.27 & \textbf{17.09} & \textbf{0.21} & 0.18 & \textbf{0.19} & \textbf{0.4373} & 58.86 \\
        & & & $m=2$ &  7.9 &   8.0 & 14.41 & 18.22 & 15.08 & 0.11 & 0.21 & 0.15 & 0.4910 & \textbf{61.65} \\
        & & & $m=3$ & 11.6 &  13.4 & 10.78 & \textbf{20.05} & 13.21 & 0.13 & \textbf{0.36} & 0.18 & 0.5632 & 51.69 \\
        \cline{3-14}
        & & \parbox[t]{2mm}{\multirow{3}{*}{\rotatebox[origin=c]{90}{pruned}}}
            & $m=1$ &  3.9 &   2.8 & \textbf{22.81} & 14.98 & \textbf{17.30} & \textbf{0.23} & 0.14 & 0.17 & \textbf{0.4282} & 60.88 \\
        & & & $m=2$ &  6.9 &   6.1 & 15.40 & 17.87 & 15.80 & 0.11 & 0.14 & 0.12 & 0.4633 & \textbf{67.48} \\
        & & & $m=3$ &  9.8 &   9.8 & 11.97 & \textbf{19.76} & 14.30 & 0.14 & \textbf{0.32} & \textbf{0.19} & 0.5272 & 58.15 \\
        
        \midrule \midrule
        
        \parbox[t]{2mm}{\multirow{12}{*}{\rotatebox[origin=c]{90}{w/o EW\textsubscript{A}}}} & \parbox[t]{2mm}{\multirow{6}{*}{\rotatebox[origin=c]{90}{w/o RT}}} 
          & & $m=1$ & 11.8 &  18.2 & \textbf{17.45} & 14.46 & \textbf{13.21} & \textbf{0.19} & 0.22 & \textbf{0.19} & \textbf{0.5193} & \textbf{45.41} \\
        & & & $m=2$ & 36.7 &  66.2 &  8.36 & 18.33 &  8.99 & 0.13 & \textbf{0.44} & 0.16 & 0.6663 & 36.46 \\
        & & & $m=3$ & 79.1 & 153.7 &  4.22 & \textbf{21.08} &  5.81 & 0.08 & \textbf{0.44} & 0.11 & 0.8078 & 19.56 \\
        \cline{3-14}
        & & \parbox[t]{2mm}{\multirow{3}{*}{\rotatebox[origin=c]{90}{pruned}}}
            & $m=1$ &  4.2 &  3.2 & \textbf{20.49} & 13.94 & \textbf{15.66} & \textbf{0.30} & 0.22 & \textbf{0.24} & \textbf{0.4340} & \textbf{59.04} \\
        & & & $m=2$ & 13.2 & 18.3 & 13.02 & 17.11 & 13.54 & 0.19 & \textbf{0.32} & 0.22 & 0.5142 & 58.31 \\
        & & & $m=3$ & 32.5 & 56.2 &  9.69 & \textbf{19.38} & 11.62 & 0.14 & \textbf{0.32} & 0.18 & 0.6215 & 44.55 \\
        \cline{2-14}
         & \parbox[t]{2mm}{\multirow{6}{*}{\rotatebox[origin=c]{90}{w/ RT}}} &
            & $m=1$ & 15.5 &  28.4 & \textbf{18.47} & 15.78 & \textbf{14.14} & \textbf{0.14} & 0.26 & \textbf{0.14} & \textbf{0.5342} & \textbf{45.22} \\
        & & & $m=2$ & 46.4 &  95.7 &  8.11 & 19.18 &  8.90 & 0.04 & 0.29 & 0.07 & 0.6948 & 32.81 \\
        & & & $m=3$ & 91.8 & 201.6 &  4.28 & \textbf{21.75} &  5.95 & 0.04 & \textbf{0.42} & 0.07 & 0.8223 & 17.52 \\
        \cline{3-14}
        & & \parbox[t]{2mm}{\multirow{3}{*}{\rotatebox[origin=c]{90}{pruned}}}
            & $m=1$ &  3.9 &  3.0 & \textbf{22.70} & 14.97 & \textbf{17.24} & \textbf{0.20} & 0.14 & \textbf{0.16} & \textbf{0.4320} & \textbf{61.29} \\
        & & & $m=2$ & 14.9 & 24.2 & 14.41 & 17.93 & 14.80 & 0.09 & 0.14 & 0.10 & 0.5079 & 59.98 \\
        & & & $m=3$ & 33.0 & 63.2 & 10.81 & \textbf{20.06} & 12.88 & 0.10 & \textbf{0.39} & 0.15 & 0.6091 & 45.91 \\
        
        \midrule \midrule
        
        \parbox[t]{2mm}{\multirow{12}{*}{\rotatebox[origin=c]{90}{w/o EW\textsubscript{O}}}} & \parbox[t]{2mm}{\multirow{6}{*}{\rotatebox[origin=c]{90}{w/o RT}}} 
          & & $m=1$ &  5.3 &  5.0 & \textbf{18.85} & 14.03 & \textbf{14.59} & \textbf{0.27} & 0.22 & \textbf{0.22} & \textbf{0.4662} & \textbf{52.38} \\
        & & & $m=2$ & 12.2 & 15.0 & 10.64 & 17.05 & 11.58 & 0.20 & \textbf{0.38} & \textbf{0.22} & 0.5686 & 48.14 \\
        & & & $m=3$ & 21.1 & 29.0 &  7.09 & \textbf{19.26} &  9.28 & 0.13 & \textbf{0.38} & 0.17 & 0.6681 & 35.71 \\
        \cline{3-14}
        & & \parbox[t]{2mm}{\multirow{3}{*}{\rotatebox[origin=c]{90}{pruned}}}
            & $m=1$ &  4.1 &  3.1 & \textbf{20.53} & 13.90 & \textbf{15.67} & \textbf{0.30} & 0.22 & 0.24 & \textbf{0.4326} & 59.21 \\
        & & & $m=2$ &  8.1 &  7.9 & 13.35 & 16.75 & 13.91 & 0.23 & \textbf{0.32} & \textbf{0.25} & 0.4895 & \textbf{61.44} \\
        & & & $m=3$ & 13.0 & 14.7 & 10.22 & \textbf{18.77} & 12.34 & 0.17 & \textbf{0.32} & 0.21 & 0.5661 & 50.95 \\
        \cline{2-14}
         & \parbox[t]{2mm}{\multirow{6}{*}{\rotatebox[origin=c]{90}{w/ RT}}}
          & & $m=1$ &  5.1 &  5.0 & \textbf{20.66} & 15.05 & \textbf{16.06} & \textbf{0.18} & 0.18 & \textbf{0.17} & \textbf{0.4625} & \textbf{54.94} \\
        & & & $m=2$ & 11.6 & 15.4 & 11.57 & 17.98 & 12.65 & 0.08 & 0.21 & 0.12 & 0.5651 & 48.40 \\
        & & & $m=3$ & 19.3 & 29.3 &  7.72 & \textbf{20.10} & 10.15 & 0.09 & \textbf{0.42} & 0.14 & 0.6656 & 35.32 \\
        \cline{3-14}
        & & \parbox[t]{2mm}{\multirow{3}{*}{\rotatebox[origin=c]{90}{pruned}}}
            & $m=1$ &  3.9 &  3.0 & \textbf{22.57} & 14.74 & \textbf{17.07} & \textbf{0.22} & 0.14 & 0.17 & \textbf{0.4315} & 61.30 \\
        & & & $m=2$ &  7.4 &  7.6 & 14.93 & 17.62 & 15.34 & 0.13 & 0.17 & 0.14 & 0.4792 & \textbf{63.78} \\
        & & & $m=3$ & 11.3 & 13.7 & 11.38 & \textbf{19.58} & 13.64 & 0.14 & \textbf{0.42} & \textbf{0.19} & 0.5544 & 52.13 \\
    \bottomrule
    \end{tabular}
    }
    \caption{Intrinsic evaluation of pruned CCKGs constructed from CN. \textit{w/o EW\textsubscript{A}} and \textit{w/o EW\textsubscript{O}} are the baselines with unweighted shortest paths described in \S \ref{sec:experiments}
    }
    \label{tab:app:eg_cn}
\end{table*}

\paragraph{Support vs. counter instances} \label{sec:app:res_eg_support_counter}
\T\ref{tab:app:eg_support_vs_counter} shows the intrinsic evaluation for \textit{support} and \textit{counter} instances separately, i.e. we split the dev set according to the gold stance label. Overall, the results are similar for support and counter instances, except for the concept precision where the supports are more than \SI{4}{\pp} better. Hence, we do not explicitly show the difference between support and counter in the rest of this paper. 

\begin{table*}
    \centering
     \resizebox{.9\linewidth}{!}{
    \begin{tabular}{cc|cccccccccc}
    \toprule
        \multicolumn{2}{c|}{Configuration} & \#nodes & \#edges & C P $\uparrow$ & C R $\uparrow$ & C F1 $\uparrow$ & T P $\uparrow$ & T R $\uparrow$ & T F1 $\uparrow$ & GED $\downarrow$ & G-BS $\uparrow$ \\
        \midrule
        \parbox[t]{2mm}{\multirow{3}{*}{\rotatebox[origin=c]{90}{all}}} 
        & $m=1$ &  4.0 &  3.0 & \textbf{52.54} & 37.94 & \textbf{42.67} & \textbf{28.55} & 19.78 & \textbf{22.13} & \textbf{0.3435} & 66.88 \\
        & $m=2$ &  6.6 &  5.8 & 36.67 & 44.36 & 38.88 & 19.42 & 25.44 & 20.97 & 0.3745 & \textbf{74.26} \\
        & $m=3$ &  9.2 &  8.5 & 29.25 & \textbf{48.55} & 35.49 & 15.51 & \textbf{29.63} & 19.56 & 0.4313 & 64.50 \\
        \midrule
        \parbox[t]{2mm}{\multirow{3}{*}{\rotatebox[origin=c]{90}{support}}} 
        & $m=1$ &  3.8 & 2.8 & \textbf{54.76} & 37.58 & \textbf{43.19} & \textbf{28.40} & 18.55 & \textbf{21.27} & \textbf{0.3511} & 64.83 \\
        & $m=2$ & 6.3 & 5.5 & 38.22 & 44.66 & 39.97 & 19.78 & 25.03 & 21.11 & 0.3744 & \textbf{74.70} \\ 
        & $m=3$ & 8.7 & 8.1 & 30.95 & \textbf{49.26} & 36.97 & 16.24 & \textbf{29.56} & 20.08 & 0.4219 & 66.20 \\
        \midrule
        \parbox[t]{2mm}{\multirow{3}{*}{\rotatebox[origin=c]{90}{counter}}} 
        & $m=1$ & 4.2 & 3.2 & \textbf{50.32} & 38.30 & \textbf{42.15} & \textbf{28.69} & 21.02 & \textbf{22.99} & \textbf{0.3359} & 68.93 \\
        & $m=2$ & 6.9 & 6.0 & 35.13 & 44.05 & 37.78 & 19.07 & 25.84 & 20.82 & 0.3746 & \textbf{73.82} \\
        & $m=3$ & 9.6 & 8.9 & 27.55 & \textbf{47.83} & 34.01 & 14.79 & \textbf{29.71} & 19.04 & 0.4407 & 62.80 \\
        \bottomrule
    \end{tabular}}
    \caption{Intrinsic evaluation of pruned CCKGs constructed from ExplaKnow on the ExplaGraphs dev split. Results are shown on i) all 398 instances, ii) the 199 \textit{support} instances and iii) the 199 \textit{counter} instances.}
    \label{tab:app:eg_support_vs_counter}
\end{table*}

\paragraph{Verbalization}
\T\ref{tab:app:eg_verbalization} shows the intrinsic evaluation for \textit{natural} and \textit{static} verbalization templates. The verbalization has a small impact on the results, but the natural verbalization yields better results overall. 

In our extrinsic evaluation the verbalization has a larger impact. This could be due to the fact that we evaluate our method extrinsically on CN instead of ExplaKnow. Due to the increased number of triplets in CN a more precise differentiation by the natural verbalization could be more important in CN than in ExplaKnow.  

\begin{table*}
    \centering
     \resizebox{.9\linewidth}{!}{
    \begin{tabular}{cc|cccccccccc}
    \toprule
        \multicolumn{2}{c|}{Configuration} & \#nodes & \#edges & C P $\uparrow$ & C R $\uparrow$ & C F1 $\uparrow$ & T P $\uparrow$ & T R $\uparrow$ & T F1 $\uparrow$ & GED $\downarrow$ & G-BS $\uparrow$ \\
        \midrule
        \parbox[t]{2mm}{\multirow{3}{*}{\rotatebox[origin=c]{90}{natural}}} 
        & $m=1$ &  4.0 &  3.0 & \textbf{52.54} & 37.94 & \textbf{42.67} & \textbf{28.55} & 19.78 & \textbf{22.13} & \textbf{0.3435} & 66.88 \\
        & $m=2$ &  6.6 &  5.8 & 36.67 & 44.36 & 38.88 & 19.42 & 25.44 & 20.97 & 0.3745 & \textbf{74.26} \\
        & $m=3$ &  9.2 &  8.5 & 29.25 & \textbf{48.55} & 35.49 & 15.51 & \textbf{29.63} & 19.56 & 0.4313 & 64.50 \\
        \midrule
        \parbox[t]{2mm}{\multirow{3}{*}{\rotatebox[origin=c]{90}{static}}}
        & $m=1$ & 3.8 & 2.8 & \textbf{52.41} & 37.07 & \textbf{42.03} & \textbf{28.06} & 19.15 & \textbf{21.51} & \textbf{0.3478} & 65.71 \\
        & $m=2$ & 6.4 & 5.5 & 36.52 & 42.94 & 38.24 & 19.25 & 24.47 & 20.52 & 0.3760 & \textbf{74.50} \\
        & $m=3$ & 8.8 & 8.1 & 28.90 & \textbf{46.73} & 34.69 & 14.94 & \textbf{27.95} & 18.68 & 0.4320 & 65.92 \\
        \bottomrule
    \end{tabular}}
    \caption{Intrinsic evaluation of pruned CCKGs constructed from ExplaKnow with \textit{natural} and \textit{static} verbalization.
    }
    \label{tab:app:eg_verbalization}
\end{table*}

\paragraph{Multiple shortest paths}
There can be potentially many contextually relevant reasoning paths between each pair of concepts. Hence, considering only the single weighted shortest path between each concept-pair might be too restrictive in the CCKG construction. 

Using Yen's algorithm we can compute the $k$ weighted shortest paths between two concepts, where $k$ is another hyperparameter. Dijkstra's algorithm can be seen as the special case of Yen's algorithm with $k=1$. 
However, using Yen's algorithm comes at increased costs for us, since Yen's algorithm only computes paths between two specific concepts, while Dijkstra's algorithm computes the shortest paths from one concept to all other concepts in one go. Thus, Yen's algorithm has to be run $m(m-1)$ times, while Dijkstra's algorithm only has to be run $m-1$ times, where $m$ is the number of initially extracted concepts. Furthermore, the time-complexity of Yen's algorithm is $kn$ times Dijkstra's algorithm's time-complexity, where $n$ is the number of concepts in the KG ($n \sim 1,000,000$ for CN). Hence, the path extraction for CCKGs with $k$ shortest paths takes $mkn$ times longer compared to our normal approach.\footnote{Code for Yen's algorithm adapted from\\\url{https://gist.github.com/ALenfant/5491853}}

\T\ref{tab:app:eg_yen_k3} shows the results for pruned CCKGs with $k=1$ and $k=3$. Without pruning, the CCKGs with $k=3$ are larger, leading to a higher recall but lower precision. Overall, the F1 score decreases as the decreased precision outweighs the increased recall. When applying pruning, $k$ only has small effects on F1 scores, with $k=1$ achieving the best performance. Hence, higher value of $k$ lead to increased computational costs without increasing performance.

\begin{table*}
    \centering
     \resizebox{.9\linewidth}{!}{
    \begin{tabular}{ccc|cccccccccc}
    \toprule
        $k$ & & $m$ & \#nodes & \#edges & C P $\uparrow$ & C R $\uparrow$ & C F1 $\uparrow$ & T P $\uparrow$ & T R $\uparrow$ & T F1 $\uparrow$ & GED $\downarrow$ & G-BS $\uparrow$ \\
        \midrule
        \multirow{6}{*}{1}
        & & 1 & 4.1 &  3.2 & \textbf{52.10} & 38.28 & \textbf{42.58} & \textbf{28.12} & 20.19 & \textbf{22.02} & \textbf{0.3458} & 66.41 \\
        & & 2 &  7.1 &  6.6 & 35.90 & 45.40 & 38.53 & 18.68 & 26.60 & 20.55 & 0.3872 & \textbf{71.39} \\
        & & 3 & 10.1 & 10.3 & 28.26 & \textbf{49.96} & 34.81 & 14.43 & \textbf{31.11} & 18.61 & 0.4524 & 60.53 \\
        \cmidrule{2-13}
        & \parbox[t]{2mm}{\multirow{3}{*}{\rotatebox[origin=c]{90}{pruned}}}
          & 1 &  4.0 &  3.0 & \textbf{52.54} & 37.94 & \textbf{42.67} & \textbf{28.55} & 19.78 & \textbf{22.13} & \textbf{0.3435} & 66.88 \\
        & & 2 &  6.6 &  5.8 & 36.67 & 44.36 & 38.88 & 19.42 & 25.44 & 20.97 & 0.3745 & \textbf{74.26} \\
        & & 3 &  9.2 &  8.5 & 29.25 & \textbf{48.55} & 35.49 & 15.51 & \textbf{29.63} & 19.56 & 0.4313 & 64.50 \\
        
        \midrule
        
        \multirow{6}{*}{3}
        & & 1 &  9.7 & 12.4 & \textbf{31.95} & 51.48 & \textbf{37.33} & \textbf{15.75} & 35.90 & \textbf{19.78} & \textbf{0.4545} & \textbf{51.03} \\
        & & 2 & 16.6 & 23.0 & 20.53 & 58.25 & 29.26 &  9.20 & 42.63 & 14.46 & 0.5924 & 33.46 \\
        & & 3 & 17.8 & 24.9 & 20.23 & \textbf{60.87} & 29.36 &  9.19 & \textbf{45.74} & 14.75 & 0.6055 & 31.68 \\
        \cmidrule{2-13}
        & \parbox[t]{2mm}{\multirow{3}{*}{\rotatebox[origin=c]{90}{pruned}}}
          & 1 & 4.0 & 3.1 & \textbf{52.38} & 38.21 & \textbf{42.68} & \textbf{28.40} & 19.96 & \textbf{22.12} & \textbf{0.3445} & 66.79 \\
        & & 2 & 6.9 & 6.2 & 36.29 & 44.79 & 38.65 & 18.96 & 26.04 & 20.71 & 0.3802 & \textbf{72.83} \\
        & & 3 & 7.7 & 7.3 & 34.76 & \textbf{47.92} & 38.84 & 18.37 & \textbf{29.42} & 21.41 & 0.3934 & 69.34 \\
        \bottomrule
    \end{tabular}}
    \caption{Intrinsic evaluation of CCKGs constructed from ExplaKnow. The number of shortest paths between each pair of extracted concepts is $k=1$ and $k=3$. 
    }
    \label{tab:app:eg_yen_k3}
\end{table*}

\paragraph{Different pruning methods}
We prune by ranking concepts according to their semantic similarity to the argument, as measured by SBERT. This reduces noise, as contextually irrelevant (i.e. dissimilar) concepts are removed. We expect that to some extent this similarity should also be reflected in the graph structure, and central concepts should be more relevant. Thus, we also try pruning by ranking concepts according to their PageRank. We recompute PageRank after each concept-deletion to ease pruning of chains of concepts. 

\T\ref{tab:app:eg_pruning} shows that the two pruning methods perform similarly; both increasing precision at the expense of a lower recall. However, pruning by SBERT shows comparable or better performance as pruning by PageRank in all metrics. Thus, we rely on SBERT for pruning. 

\begin{table*}
\centering
     \resizebox{.9\linewidth}{!}{
    \begin{tabular}{cc|rrrrrrrrrr}
    \toprule
        \multicolumn{2}{c|}{Configuration} & \#nodes & \#edges & C P $\uparrow$ & C R $\uparrow$ & C F1 $\uparrow$ & T P $\uparrow$ & T R $\uparrow$ & T F1 $\uparrow$ & GED $\downarrow$ & G-BS $\uparrow$ \\
        
        \midrule
        
        \parbox[t]{2mm}{\multirow{3}{*}{\rotatebox[origin=c]{90}{None}}} 
        & $m=1$ &  4.1 &  3.2 & \textbf{52.10} & 38.28 & \textbf{42.58} & \textbf{28.12} & 20.19 & \textbf{22.02} & \textbf{0.3458} & 66.41 \\
        & $m=2$ &  7.1 &  6.6 & 35.90 & 45.40 & 38.53 & 18.68 & 26.60 & 20.55 & 0.3872 & \textbf{71.39} \\
        & $m=3$ & 10.1 & 10.3 & 28.26 & \textbf{49.96} & 34.81 & 14.43 & \textbf{31.11} & 18.61 & 0.4524 & 60.53 \\
        
        \midrule
        
        \parbox[t]{2mm}{\multirow{3}{*}{\rotatebox[origin=c]{90}{SB}}}
        & $m=1$ & 4.0 & 3.0 & \textbf{52.54} & 37.94 & \textbf{42.67} & \textbf{28.55} & 19.78 & \textbf{22.13} & \textbf{0.3435} & 66.88 \\
        & $m=2$ & 6.6 & 5.8 & 36.67 & 44.36 & 38.88 & 19.42 & 25.44 & 20.97 & 0.3745 & \textbf{74.26} \\
        & $m=3$ & 9.2 & 8.5 & 29.25 & \textbf{48.55} & 35.49 & 15.51 & \textbf{29.63} & 19.56 & 0.4313 & 64.50 \\
        
        \midrule
        
        \parbox[t]{2mm}{\multirow{3}{*}{\rotatebox[origin=c]{90}{PR}}}
        & $m=1$ & 4.0 & 3.0 & \textbf{52.26} & 37.58 & \textbf{42.36} & \textbf{28.14} & 19.30 & \textbf{21.70} & \textbf{0.3445} & 66.83 \\
        & $m=2$ & 6.6 & 5.7 & 36.38 & 43.71 & 38.48 & 18.93 & 24.47 & 20.34 & 0.3763 & \textbf{74.23} \\
        & $m=3$ & 9.1 & 8.5 & 29.04 & \textbf{48.03} & 35.20 & 15.17 & \textbf{28.86} & 19.09 & 0.4327 & 64.46 \\

    \bottomrule
    \end{tabular}
    }
    \caption{Intrinsic evaluation of different pruning methods on ExplaKnow. Pruning is ranked by \textit{None}: no pruning; \textit{SB}: SBERT; \textit{PR}: PageRank. %
    }
    \label{tab:app:eg_pruning}
\end{table*}

\paragraph{Constituent parser for concept extraction}
In the extrinsic evaluation (\S \ref{sec:extrinsic}), we face the problem that arguments consist of long premises and conclusions. Extracting concepts with our usual approach yields concepts that match the premise and argument, but often they do \textit{not} cover all aspects of the text. Hence, we first parse the texts into \textit{constituents}, and then extract concepts for each constituent individually. Please refer to \S \ref{sec:app:valnov_modelvariations} for more details. 

\T\ref{app:tab:intrinsic_concept_extraction} shows the results when relying on constituents for concept extraction. Using the constituents more than doubles the CCKGs in size, but also increases concept and triplet recall by more than \SI{30}{\pp} The precision on the other hand decreases due to the increased graph size. Overall the concept F1 scores decrease and the triplet F1 scores increase slightly. However, the structural similarity to the gold graphs, as measured by GED and G-BS, decreases as a result of the larger graph sizes. Thus, not using constituents for concept extraction achieves better scores overall in the intrinsic evaluation. We expect that this would change in an evaluation with longer sentences and larger gold graphs. 

\begin{table*}
    \centering
     \resizebox{.9\linewidth}{!}{
    \begin{tabular}{ccc|rrrrrrrrrr}
    \toprule
        \multicolumn{3}{c|}{Configuration} & \#nodes & \#edges & C P $\uparrow$ & C R $\uparrow$ & C F1 $\uparrow$ & T P $\uparrow$ & T R $\uparrow$ & T F1 $\uparrow$ & GED $\downarrow$ & G-BS $\uparrow$ \\
        \midrule
        \parbox[t]{2mm}{\multirow{10}{*}{\rotatebox[origin=c]{90}{w/o constituents}}} 
       & & $m=1$ &  4.1 &   3.2 & \textbf{52.10} & 38.28 & \textbf{42.58} & \textbf{28.12} & 20.19 & \textbf{22.02} & \textbf{0.3458} & 66.41 \\
       & & $m=2$ &  7.1 &   6.6 & 35.90 & 45.40 & 38.53 & 18.68 & 26.60 & 20.55 & 0.3872 & \textbf{71.39} \\
       & & $m=3$ & 10.1 &  10.3 & 28.26 & 49.96 & 34.81 & 14.43 & 31.11 & 18.61 & 0.4524 & 60.53 \\
       & & $m=4$ & 13.0 &  14.1 & 23.26 & 52.37 & 31.11 & 11.59 & 33.45 & 16.30 & 0.5158 & 50.47 \\
       & & $m=5$ & 15.9 &  17.8 & 19.94 & \textbf{54.41} & 28.28 &  9.88 & \textbf{36.08} & 14.81 & 0.5637 & 42.98 \\
        
        \cmidrule{2-13}
        
        & \parbox[t]{2mm}{\multirow{5}{*}{\rotatebox[origin=c]{90}{pruned}}}
         & $m=1$ &  4.0 &  3.0 & \textbf{52.54} & 37.94 & \textbf{42.67} & \textbf{28.55} & 19.78 & \textbf{22.13} & \textbf{0.3435} & 66.88 \\
       & & $m=2$ &  6.6 &  5.8 & 36.67 & 44.36 & 38.88 & 19.42 & 25.44 & 20.97 & 0.3745 & \textbf{74.26} \\
       & & $m=3$ &  9.2 &  8.5 & 29.25 & 48.55 & 35.49 & 15.51 & 29.63 & 19.56 & 0.4313 & 64.50 \\
       & & $m=4$ & 11.5 & 11.2 & 24.43 & 50.94 & 32.17 & 12.77 & 31.88 & 17.58 & 0.4890 & 54.90 \\
       & & $m=5$ & 13.8 & 13.8 & 21.17 & \textbf{52.87} & 29.54 & 10.96 & \textbf{34.14} & 16.11 & 0.5337 & 47.59 \\
        
        \midrule
        
        \parbox[t]{2mm}{\multirow{10}{*}{\rotatebox[origin=c]{90}{w/ constituents}}}
        & & $m=1$ & 19.5 &  24.4 & \textbf{26.47} & 71.13 & \textbf{36.30} & \textbf{16.13} & 55.71 & \textbf{23.00} & \textbf{0.5489} & \textbf{39.86} \\
        & & $m=2$  & 36.5 &  51.9 & 15.97 & 78.63 & 25.32 &  8.91 & 64.71 & 14.84 & 0.7038 & 22.40 \\
        & & $m=3$  & 52.9 &  80.0 & 11.25 & 82.79 & 19.16 &  5.83 & 69.72 & 10.40 & 0.7843 & 15.01 \\
        & & $m=4$  & 68.8 & 109.0 &  8.91 & 85.12 & 15.67 &  4.51 & 72.88 &  8.25 & 0.8286 & 11.42 \\
        & & $m=5$  & 85.2 & 139.7 &  7.31 & \textbf{87.23} & 13.17 &  3.60 & \textbf{75.70} &  6.71 & 0.8600 & 8.99 \\
        
        \cmidrule{2-13}
        
        & \parbox[t]{2mm}{\multirow{5}{*}{\rotatebox[origin=c]{90}{pruned}}}
          & $m=1$ & 13.8 & 13.7 & \textbf{30.53} & 66.47 & \textbf{40.18} & \textbf{19.63} & 48.57 & \textbf{26.51} & \textbf{0.4690} & \textbf{51.51} \\
        & & $m=2$ & 23.4 & 24.9 & 20.29 & 74.41 & 30.91 & 12.62 & 57.35 & 19.99 & 0.6076 & 33.72 \\
        & & $m=3$ & 32.9 & 37.2 & 15.27 & 78.90 & 24.97 &  9.30 & 62.98 & 15.79 & 0.6895 & 24.89 \\
        & & $m=4$ & 42.2 & 50.3 & 12.43 & 81.44 & 21.10 &  7.49 & 66.41 & 13.15 & 0.7406 & 19.94 \\
        & & $m=5$ & 51.4 & 63.1 & 10.56 & \textbf{83.91} & 18.40 &  6.34 & \textbf{69.96} & 11.41 & 0.7777 & 16.45 \\
        \bottomrule
    \end{tabular}
    }
    \caption{Intrinsic evaluation of different concept extractions for pruned CCKGs constructed from ExplaKnow. 
    }
    \label{app:tab:intrinsic_concept_extraction}
\end{table*}

\subsection{ExplaGraphs manual evaluation}\label{sec:app:eg_manual_detailed}
\subsubsection{Annotation description} \label{sec:app:eg_manual_detailed_description}
For each instance, we asked a series of questions to annotators for which they had to select one answer or say that they can not make a decision. The first set of questions revolved around the argument as such, without considering the graph. In \textbf{Q1} annotators selected the correct of 9 predefined \textit{topics}. Next, in \textbf{Q2}, we asked whether the conclusion is \textit{plausible} given the premise. We asked this to assess i) quality of \citet{saha-etal-2021-explagraphs}'s arguments, and ii) whether we obtain plausible premise-conclusion pairs from the belief-argument pairs. If an argument was labeled as plausible, then in \textbf{Q3} annotators had to decide if they can identify an \textit{implicit CSK connection} that links the conclusion to the premise. If so, we also ask the annotators to formulate and write down the perceived CSK connection in plain language. This serves to familiarize the annotators with the argument, and provides them with a reference to their own interpretation in the later graph quality assessment steps. 

The second set of questions were only presented for plausible arguments with a perceived CSK connection, to assess the quality of the provided CCKGs. \textbf{Q4}: To estimate the \textit{recall} we asked if the graph shows the implicit connection i) \textit{completely} ii) \textit{partially} or iii) \textit{not at all}. Then, to estimate \textit{precision} at a fine-grained analysis level, each individual triplet had to be labeled in \textbf{Q5} as i) \textit{positive} (expresses implicit CSK) ii) \textit{neutral} (does not express implicit CSK, but matches the topic) iii) \textit{unrelated} (does not match the topic) or iv) \textit{negative} (contradicts implicit CSK or the conclusion, but the topic is appropriate). An example of a \textit{negative} triplet would be $\triple{human\_cloning}{IsA}{considered\_unethical}$ in a CCKG for an argument with a pro-cloning conclusion. 
For \textit{negative} triplets, we further asked (\textbf{Q6}) if its \textit{negation} expresses relevant implicit CSK, and (\textbf{Q7}) if the graph extended with the negated triplet(s) shows the CSK connection. However, \textit{negative} triplets were rare in our CCKGs, such that we could not perform analysis of Q6 and Q7. 

Please refer to our official annotation guidelines at \url{https://github.com/Heidelberg-NLP/CCKG/blob/main/annotation_guidlines.pdf} for more details on each question, as well as illustrative examples. 

\subsubsection{Annotation results} \label{sec:app:eg_manual_detailed_results}
For each question, \T\ref{tab:app:eg_manual_detailed_results} reports the \textit{support}, i.e. the number of instances that were annotated by both annotators. Note that the support decreases in Q3 and Q4, since annotation instances that were labeled with \textit{no} in Q2 (\textit{plausible argument?}) or Q3 (\textit{implicit CSK in argument?}) were not further annotated by the individual annotators. We only report values for which both annotators provided labels. Q5 has a support of 1,169 triplets that come from the same 115 graphs as annotated in Q4. 

To measure \textbf{inter-annotator agreement}, we report the counts of the assigned labels per class and annotator (\textit{A1}, \textit{A2}), and compute agreement scores using a) \textit{Cohen's Kappa} $\kappa$, where we compute $\kappa$ of individual labels in a one-vs-all setting, i.e. by considering all other labels as the same label. This we complement by b) counts and percentages of the \textit{overlap of label assignments} (A1 $\wedge$ A2) by the two annotators per class.\footnote{The percentage is computed relative to the average of A1 and A2.} %
We also report the percentage of labels assigned by both annotators unanimously or by at least one annotator. %

We now investigate the annotation results on Q1 to Q5. 

\textbf{Q1} (\textit{Topic}): The arguments are uniformly distributed across topics. The topics are quite distinct such that the annotators could assign them to the correct classes with ease, with only minimal divergences, yielding a high inter-annotator agreement ($\kappa = 0.916$). 

\textbf{Q2} (\textit{plausible?}): A large majority of instances (\SI{79.90}{\%}) were unanimously labeled as plausible, which shows that \citet{saha-etal-2021-explagraphs}'s support \textit{belief}-\textit{argument} pairs can indeed be interpreted as \textit{premise}-\textit{conclusion} pairs. 

However, $\kappa$ is low, as one annotator considered  all but 3 arguments as plausible, while the other considered 38 of the 199 arguments, i.e., \SI{19}{\%}, as implausible.
On deeper inspection we found that these \SI{19}{\%} suffered from various deficiencies: multiple negations made interpretation very difficult and did often not yield a valid supporting argument; in other cases the pairs were presented in the wrong direction to count as an argument. One of our annotators considered the arguments with great care and we could validate his judgements in almost all cases. We are therefore confident that the vast majority of such cases could be captured in our annotation.

\textbf{Q3} (\textit{implicit CSK in argument?)}: Only \SI{6.29}{\%} of arguments were unanimously judged as not being linked through implicit CSK, which confirms that \citet{saha-etal-2021-explagraphs}'s data collection successfully resulted in \textit{belief-argument} pairs that require explanations. In \SI{72.33}{\%} of cases both annotators agreed that there is implicit CSK (115 instances). On these 115 instances we evaluate the performance of our CCKGs. 

\textbf{Q4} (\textit{CSK covered in CCKG?}): Here the annotators evaluated whether the presented CCKG covered the implicit knowledge, by referring to what they had written down in Q3, but they could also accept another valid interpretation expressed by the graph. \SI{29.57}{\%} of CCKGs were unanimously annotated to cover the implicit CSK \textit{completely}, i.e. the argument could be fully understood based on knowledge shown in the CCKG. When considering CCKGs annotated by at least one annotator as complete, the score doubles to \SI{59.13}{\%}. \SI{88.70}{\%} were unanimously judged to cover the implicit CSK at least \textit{partially}, which corresponds to a \textit{high recall of implicit CSK} in the constructed CCKGs. I.e., most CCKGs make the connection between conclusion and premise more explicit, and hence, they can be expected to support computational systems in knowledge-intense argumentation tasks. With 0.413, Cohen's~$\kappa$ is higher for \textit{completely} than for \textit{partially}, indicating that partial coverage is more subjective to decide. 

\textbf{Q5} (\textit{Triplet rating}): The remaining 115 CCKGs contain 1,169 triplets in total. Out of these, \SI{39.44}{\%} were unanimously labeled as \textit{positive}, i.e., the triplet \textit{reflects implicit CSK} that links the conclusion to the premise (again, annotators are asked to compare the CCKG to their answer to Q3, but are free to accept other valid connections in the CCKG), and for \SI{74.68}{\%} at least one annotator rated the triplet as positive. This shows that a substantial amount of triplets reflect implicit CSK, while the judgement may be subjective, depending on the annotator's own interpretation. Also, it is often difficult to decide what the exact implicit CSK is. 

\SI{13.94}{\%} of all triplets were unanimously labeled \textit{neutral}, i.e. they express knowledge pertaining to the topic of the argument. As such, they contribute additional knowledge or context for the argument, but no CSK that is required to support the conclusion.

Only \SI{1.71}{\%} of triples were unanimously labeled as \textit{unrelated}, i.e. as not matching the argument because they no not match the topic. These triplets represent noise in the CCKG, and are mostly avoided by the strong contextualization during graph construction. Only a small number remains after pruning. 

\SI{1.07}{\%} of all triplets are unanimously labeled \textit{negative}, i.e. they contradict the conclusion or the implicit CSK. These triplets are from the correct topic, but often show the issue from a different perspective and do not support the conclusion.

In the first block of \T\ref{tab:app:eg_manual_detailed_results_macro}, we also report macro averages over the triplet precision measured in Q5 (triplet rating) for individual graphs. We report the score for triplets showing implicit CSK (i.e. \textit{positive} triplets) and triplets being from the correct topic (i.e. \textit{positive}, \textit{neutral} or \textit{negative} triplets). Again, we report the \textit{support} and the values for each individual annotator \textit{A1} and \textit{A2}. We derive a joint rating from both annotators by either i) A1 $\wedge$ A2: A triplet is only considered as positive / in-topic if both annotators labeled it as such, or ii) A1 $\vee$ A2: A triplet is considered as positive / in-topic if at least one annotator labeled it as such. 

The unanimous macro precision is \SI{39.43}{\%} for triplets showing implicit CSK and \SI{73.87}{\%} when considering triplets rated as positive by at least one annotator. This matches our observation from the micro scores. Our CCKGs show high in-topic macro precision with \SI{92.76}{\%} in the unanimous setting and exceeding \SI{99}{\%} when considering triplets rated by at least one annotator as in-topic. 

\T\ref{tab:app:eg_manual_detailed_results_macro} also shows the macro precision for graphs which were unanimously judged to reflect the implicit CSK in the argument \textit{completely} and \textit{partially} in Q4. The precision of unanimous positive triplets increases by more than \SI{15}{\pp} when considering only CCKGs that reflect the implicit CSK completely. On the other hand, the precision of in-topic triplets increases more when considering CCKGs that reflect the implicit CSK only partially. This indicates that CCKGs that fail to reflect implicit CSK completely still reflect CSK from the correct topic.

Overall, the manual annotation shows strong performance of the CCKGs in terms of implicit CSK recall, implicit CSK precision, and in-topic precision. 

\begin{table*} %
\centering
\resizebox{\linewidth}{!}{
\begin{tabular}{ll|cccc|cc|cc}
\toprule
 \multirow{2}{*}{Question} & \multirow{2}{*}{Label} & \multicolumn{4}{c|}{Counts [\#]} & \multicolumn{2}{c|}{Agreement} &\multicolumn{2}{c}{Quality [\%]} \\
& & Support & A1 & A2 & A1 $\wedge$ A2 & $\kappa$ & A1 $\wedge$ A2 [\%] & A1 $\wedge$ A2 & A1 $\vee$ A2 \\
\midrule
Q1 & all labels & 199 & & & 184 & 0.916 & \\
which   & abandon marriage & & 24 & 26 & 24 & 0.954 & 96.00 & 12.06 & 13.07 \\
topic?     & ban cosmetic surgery && 22 & 20 & 20 & 0.947 & 95.24 & 10.05 & 11.06 \\
   & adopt an austerity regime & & 22 & 20 & 19 & 0.894 & 90.48 & 9.55 & 11.56 \\
   & fight urbanization & & 22 & 22 & 22 & 1.000 & 100.00 & 11.06 & 11.06 \\
   & subsidize embryonic stem cell research & & 19 & 18 & 17 & 0.911 & 91.89 & 8.54 & 10.05 \\
   & legalize entrapment & & 22 & 22 & 22 & 1.000 & 100.00 & 11.06 & 11.06 \\
   & ban human cloning & & 21 & 21 & 19 & 0.894 & 90.48 & 9.55 & 11.56 \\
   & close Guantanamo Bay detention camp & & 21 & 21 & 21 & 1.000 & 100.00 & 10.55 & 10.55 \\
   & adopt atheism & & 22 & 25 & 19 & 0.783 & 80.85 & 9.55 & 14.07 \\
   & $\times$          & & 4 & 4 & 1 & 0.235 & 25.00 & 0.50 & 3.52 \\
\midrule
Q2 & all labels & 199 & & & 160 & 0.021 & \\
plausible   & yes & & 196 & 161 & 159 & 0.021 & 89.08 & 79.90 & 99.50 \\
argument?   & no & & 3 & 38 & 1 & 0.021 & 4.88 & 0.50 & 20.10 \\
\midrule
Q3 & all labels & 159 & & & 125 & 0.298 & \\
implicit CSK     & yes  & & 149 & 115 & 115 & 0.298 & 87.12 & 72.33 & 93.71 \\
in argument?     & no & & 10 & 44 & 10 & 0.298 & 37.04 & 6.29 & 27.67 \\
\midrule
Q4 & all labels & 115 & & & 68 & 0.268 & \\
CSK in   & completely & & 43 & 59 & 34 & 0.413 & 66.67 & 29.57 & 59.13 \\
 CCKG?   & partially &  & 59 & 56 & 34 & 0.183 & 59.13 & 29.57 & 70.43 \\
 & completely or partially & & 102 & 115 & 102 & 0.000 & 94.01 & 88.70 & 100.00 \\
   & not at all & & 13 & 0 & 0 & 0.000 & 0.00 & 0.00 & 11.30 \\
\midrule
Q5 (micro) & all labels & $\star\,$1169 & & & 656 & 0.230 & \\
triplet rating & positive   & & 556 & 778 & 461 & 0.306 & 69.12 & 39.44 & 74.68 \\
   & neutral    & & 465 & 321 & 163 & 0.133 & 41.48 & 13.94 & 53.29 \\
   & unrelated  & & 100 & 54 & 20 & 0.212 & 25.97 & 1.71 & 11.46 \\
   & negative   & & 48 & 16 & 12 & 0.362 & 37.50 & 1.03 & 4.45 \\
   & positive or neutral & & 1021 & 1099 & 985 & 0.251 & 92.92 & 84.26 & 97.09 \\
   & in-topic (i.e. all but unrelated) & & 1069 & 1115 & 1035 & 0.212 & 94.78 & 88.54 & 98.29 \\
\bottomrule
\end{tabular}}
\caption{Results of manual annotation. \textit{A1} and \textit{A2} show the label counts for each individual annotator. \textit{A1 $\wedge$ A2} shows the counts (\#) and percentages (\%) of instances labeled unanimously by \textit{A1} and \textit{A2}; \textit{A1 $\vee$ A2} shows the percentage of instance labels assigned by at least one of the annotators.  
$\times$ means that the annotator could not decide. Topic labels for Q0 are posed as "We should ..." . \\
$\star$: the supporting 1,169 triplets are from the 115 supporting CCKG graphs from Q4. }
\label{tab:app:eg_manual_detailed_results}
\end{table*}

\begin{table*} %
\centering
\resizebox{.95\linewidth}{!}{
\begin{tabular}{ll|rcccc}
\toprule
CSK shown in CCKG & Label & support & A1 & A2  & A1 $\wedge$ A2 & A1 $\vee$ A2 \\
\midrule
\multirow{2}{*}{All} & Implicit CSK & 115 & 48.36 & 64.95 & 39.43 &  73.87 \\
 & Topic &  115 & 94.94 & 97.01 & 92.76 & 99.20 \\
\midrule
\multirow{2}{*}{Completely} & Implicit CSK & 34 & 66.36 & 74.90 & 56.37 & 84.89 \\
 & Topic & 34 & 95.94 & 97.16 & 93.57 & 99.53 \\
\midrule
\multirow{2}{*}{Partial} & Implicit CSK & 34 & 48.02 & 58.39 & 36.77 & 69.64 \\
 & Topic &  34 & 99.18 & 97.79 & 97.42 & 99.55 \\
\bottomrule
\end{tabular}
}
\caption{Macro precision scores of manual annotation on Q5 (triplet rating) in \%. \textit{A1} and \textit{A2} are the macro averages for each individual annotator, $A1 \wedge A2$ is the macro average when only considering unanimous decisions and $A1 \vee A2$ is the macro average when considering triplets which at least one annotator judged as positive / in-topic. }
\label{tab:app:eg_manual_detailed_results_macro}
\end{table*}

\subsection{\valnov}
\subsubsection{Data statistics} \label{sec:app:vn_datastatistics}
\citet{heinisch-etal-2022-overview} collect arguments from diverse topics, where the conclusions are partially automatically generated. The binary labels for validity and novelty are manually created by multiple annotators.
The data for the \valnov~Shared Task\footnote{The task was organized as part of the ArgMining workshop 2022.} has been constructed from arguments from an argumentative dataset \citep{ajjour-etal-2019-modeling}, and has been extended by conclusions automatically generated with T5 \citep{heinisch-etal-2022-strategies,raffel-etal-2020-exploring}, producing instances of paired premise-conclusion pairs. All instances were manually assigned binary labels for \textit{validity} and \textit{novelty}. %

The \valnov~train/ dev/ test sets consist of 750/ 202/ 520 instances. However, 48 of the train instances are \textit{defeasible}, i.e. instances with no annotator majority for validity or novelty. We remove these instances,
leaving us with 702 training items.

The train set is unbalanced, with only \SI{2}{\%} of the train data being from the \textit{non-valid and novel} class. %

\citet{heinisch-etal-2022-data} extend the dataset by integrating datasets from different tasks as well as synthetic data. In this work we only use the original dataset proposed by \citet{heinisch-etal-2022-overview}. 

\subsubsection{Model variations} \label{sec:app:valnov_modelvariations}
\paragraph{Concept extraction with constituents}
The arguments in the \valnov~dataset are relatively long (76 tokens in avg.), often containing more that one aspect / perspective. This negatively effects the quality of triplet selection for concept extraction: the extracted concepts are semantically relevant, but often do not span the entire argument. We thus split the text into constituents using a SOTA parser \citep{zhang-etal-2020-fast}, 
and select concepts for each constituent separately. The hyperparameter $m$ now controls the number of extracted triplets for each constituent. We use their \texttt{crf-con-roberta-en} model at \url{www.github.com/yzhangcs/parser}. Leaf nodes often consist of only one or two concepts, which limits contextualization for these constituents. Hence, we disregard the leaf nodes to reduce noise in concept extraction. 

\paragraph{Partial pruning}
Pruning CCKGs completely bears the risk of removing relevant structure. However, not pruning at all leaves the CCKGs in a potentially noisy state. To allow for a more fine-grained balance, we apply \textit{partial pruning}. I.e., we rank concepts and prune the CCKG accordingly, but instead of pruning all possible concepts we only remove the first \SI{25}{\%}, \SI{50}{\%} or \SI{75}{\%}, which corresponds to removing only the most dissimilar concepts.

\subsubsection{Feature extraction} \label{sec:app:valnov_featureextraction}
\paragraph{Structural features} We extract 5 features describing the \textbf{size} of CCKGs (number of concepts, number of triplets, number of premise-concepts, number of conclusion-concepts, number of concepts shared by premise and conclusion), 6 features describing the \textbf{connectivity}  of CCKGs (number of cluster with and without edge weights and the corresponding modularity, density, transitivity), and 4 features describing the \textbf{distance} between premise and conclusion in the CCKG (weighted and unweighted MinCut between premise-concepts and conclusion-concepts, average and maximal weighted length between premise-concepts and conclusion concepts). This yields 15 graph features in total. 

\paragraph{Textual features}
We consider the \textbf{semantic similarity} between premise and conclusion (measured by SBERT), and the \textbf{NLI} probabilities that the premise \textit{entails}, is \textit{neutral} or \textit{contradicts} the conclusion. We compute the NLI predictions from a RoBERTa-large \citep{zhuang-etal-2021-robustly} model which was fine-tuned on NLI data.\footnote{We applied \texttt{roberta.large.mnli} from \\ \url{https://github.com/facebookresearch/fairseq/blob/main/examples/roberta/README.md}} This yields 4 text features in total.

\subsubsection{Classifier} \label{sec:app:valnov_classifiers}
We use scikit-learn \citep{pedregosa-etal-2011-scikit}'s \textit{RandomForest} and \textit{SVM}. For the SVM we test linear and RBF kernels. 

Our RandomForests consist of 1000 trees with Gini impurity and $4$ features considered at each split. Data is sampled with bootstrapping. For regularization we use Minimal Cost-Complexity Pruning with the hyperpaprameter $\alpha$. We choose the best value for $\alpha$ on the dev split from $\{\num{0}, \num{1e-4}, \num{5e-4}, \num{1e-3}, \num{5e-3}, \num{1e-2}, \num{5e-2}, \num{1e-1}, \num{5e-1}\}$. 

For the SMVs we apply a shrinking heuristic and choose the regularization parameter $C$ on the dev split from $\{\num{1e-5}, \num{1e-4}, \num{1e-3}, \num{1e-2}, \num{1e-1}, \num{0.5}, \num{1}, \num{2}, \num{5}, \num{10}\}$. For the RBF kernel we set $\gamma$ to \textit{scale} or \textit{auto}, also determined on the dev split. 

The best setting for CCKGs was RandomForest with $\alpha = 0.01$. For methods from \citet{saha-etal-2022-explanation} the best methods were T5: RandomForest with $\alpha=0.05$; max-margin: RandomForest with $\alpha=0.05$; and contrastive: SVM with RBF with $\gamma=auto$ and $C=5$. 

\subsubsection{Ablation} \label{sec:app:valnov_ablation}
\begin{table}
    \centering
    \resizebox{.9\linewidth}{!}{
    \begin{tabular}{cl|rrr}
    \toprule
        & configuration & joint F1 & Val F1 & Nov F1 \\
        \midrule
        & CCKG & 43.91 & 70.69 & 63.30 \\
        \midrule    
        \parbox[t]{2mm}{\multirow{6}{*}{\rotatebox[origin=c]{90}{CCKG}}}
        & w/o EW\textsubscript{O} & \textbf{-6.51} & -3.80 & -1.69 \\
        & w/o EW\textsubscript{A} & -3.25 & \textbf{-4.45} & 1.76 \\
        & w/ static verbalization & -4.32 & \textbf{-6.34} & 2.44 \\
        & w/ RelatedTo & -3.66 & \textbf{-6.86} & 2.20 \\
        & w/o pruning & \textbf{-8.24} & -2.81 & -0.14\\
        & w/ full pruning & \textbf{-5.28} & -3.39 & -1.53 \\
        \midrule
        \parbox[t]{2mm}{\multirow{6}{*}{\rotatebox[origin=c]{90}{Concept Extract.}}}
        & $m=1$  w/ const. & \textbf{-5.39} & -5.27 & -1.43 \\
        & $m=3$  w/ const. & \textbf{-7.57} & -3.03 & 0.01 \\
        & $m=1$ w/o const. & -8.11 & -3.53 & \textbf{-11.08} \\
        & $m=2$  w/o const. & -3.36 & \textbf{-5.86} & 0.79\\
        & $m=3$  w/o const. & -3.88 & \textbf{-5.83} & 0.41 \\
        & string matching & \textbf{-6.71} & -3.23 & 0.55 \\
        \midrule
        \parbox[t]{2mm}{\multirow{6}{*}{\rotatebox[origin=c]{90}{Features}}} 
        & w/o Text feats. & -20.65 & \textbf{-20.74} & -17.69 \\
        & w/o Graph feats. & \textbf{-11.65} & -3.40 & -5.12 \\
        & w/o connectivity feats. & \textbf{-5.60} & -4.01 & -0.60 \\
        & w/o size & -2.60 & \textbf{-2.93} & 0.80\\
        & w/o PC-distance feats. & -2.27 & -0.27 & \textbf{-3.73} \\
        & w/o upsampling & \textbf{-4.11} & -2.43 & -3.10 \\
        \bottomrule
    \end{tabular}}
    \caption{System ablations: values show performance differences to our full system results. 1st block: different CCKG constructions; 2nd block: configurations for concept extraction; 3rd block: CCKGs with different features / upsampling turned off.
    }
    \label{tab:valnov_ablation}
\end{table}

Our white-box feature-based system allows for a thorough ablation study (see \T\ref{tab:valnov_ablation}). We first explore variations in \textbf{CCKG construction}. 
Ablating all \textit{edge weights} incurs considerable performance losses for the joint and validity scores. Considering only one random path between each pair of concepts (w/o EW\textsubscript{O}) additionally has reduced performance for novelty. However, considering all unweighted shortest path (w/o EW\textsubscript{A}) increases the novelty score by \SI{1.76}{\pp} This indicates that contextualization is more relevant for validity, perhaps because without edge weights the model can not distinguish between valid and non-valid connections. 
The static \textit{verbalization} (see \S \ref{sec:app:method_verbalization}) reduces the quality of edge weights, and hence decreases validity score by \SI{6.34}{\pp} On the other hand, it increases the novelty score but not enough to compensate for the reduction in validity. Unspecific $\mathrm{RelatedTo}$ edges have a strong negative impact for validity but improve novelty, by attracting more knowledge. No \textit{pruning} fails to distinguish valid from non-valid conclusions due to too many noisy connections. Too much pruning on the other hand removes structural diversity and hence decreases the predictive power of CCKGs. The results suggest that contextualized graph construction has a strong impact on \textit{validity} and the \textit{joint} score, which intuitively makes sense as the contextualization promotes valid connections. At the same time, the fluctuating effects for \textit{novelty} indicate that novelty and validity are difficult to calibrate, but at a relatively low impact level. 

The impact of \textbf{concept extraction} can be best observed when comparing $m=1$ with $m=2,3$ without the constituent parser. Choosing $m=1$ results in small graphs, which can not cover all aspects of the argument. Hence, the resulting graphs are not suitable for predicting novelty. Increasing $m$ alleviates this problem, but decreases validity. We found $m=2$ with constituent parsing to yield best results. 

\textbf{Feature ablation} shows that both, \textit{text and graph features}, are necessary to achieve good performance. %
The textual features have a stronger impact on validity, while the graph features are more impactful for novelty prediction. Yet, both metrics benefit from both modalities. This indicates that text and CCKG contain complementary information and should be considered jointly in future work. Finally, we remove selected graph features from the classifier, i.e. all \textit{size}, \textit{connectivity} or \textit{premise-conclusion distance} features, at a time. This induces losses of \SI{5.60}{\pp} / \SI{4.01}{\pp} joint / validity score, for connectivity features, and strong losses of \SI{3.73}{\pp} for novelty when removing PC-distance features. This supports our hypothesis that validity correlates with the connectivity, and novelty with the distance between premise-concepts and conclusion-concepts in the CCKGs. 

\T\ref{tab:valnov_EGgraphs_ablation} shows feature ablations when constructing graphs with the \textbf{supervised methods} from \citet{saha-etal-2022-explanation}. The graph contributes more to novelty prediction in all three methods. This is consistent with previous findings, as the models leverage structural data which was found to be important for novelty. However, the effect of ablating features varies for each method and no clear trend is apparent.  
\begin{table}
    \centering
    \resizebox{.9\linewidth}{!}{
    \begin{tabular}{l|rrr}
    \toprule
        configuration & joint F1 & Val F1 & Nov F1 \\
        \midrule 
        T5 & 37.71 & 67.07 & 63.52 \\ 
        $\quad$w/o Text feats. & -10.97 & \textbf{-16.35} & -12.84 \\
        $\quad$w/o Graph feats. & \textbf{-5.45} & 0.21 & -5.34 \\
        $\quad$w/o connectivity feats. & \textbf{-0.63} & 0.26 & -0.57 \\
        $\quad$w/o size & 0.87 & -0.66 & \textbf{-1.72} \\
        $\quad$w/o PC-distance feats. & \textbf{0.64} & 0.55 & -0.08 \\
        $\quad$w/o upsampling & -0.10 & 0.49 & \textbf{-4.33}\\
        \midrule 
        max-margin & 36.22 & 67.61 & 63.27 \\ 
        $\quad$w/o Text feats. & -15.22 & -18.59 & \textbf{-19.90} \\
        $\quad$w/o Graph feats. & -3.96 & -0.33 & \textbf{-5.08} \\
        $\quad$w/o connectivity feats. & \textbf{0.67} & 0.04 & 0.52 \\
        $\quad$w/o size & \textbf{0.69} & 0.04 & 0.55 \\
        $\quad$w/o PC-distance feats. & \textbf{5.32} & -1.07 & 0.37 \\
        $\quad$w/o upsampling & -4.05 & 0.43 & \textbf{-16.79} \\
        \midrule
        contrastive & 37.82 & 64.77 & 59.96 \\
        $\quad$w/o Text feats. & -5.35 & \textbf{-10.27} & -2.70 \\
        $\quad$w/o Graph feats. & \textbf{-5.56} & 2.51 & -1.77 \\        
        $\quad$w/o connectivity feats. & -0.75 & \textbf{2.85} & -0.76 \\
        $\quad$w/o size & \textbf{-5.85} & 2.51 & -1.91 \\
        $\quad$w/o PC-distance feats. & -1.00 & \textbf{4.28} & -3.49 \\
        $\quad$w/o upsampling & 0.39 & \textbf{-0.44} & 0.27 \\
        \bottomrule
    \end{tabular}}
    \caption{Feature ablations for \citet{saha-etal-2022-explanation}'~graphs with our feature extraction and classification. Ablated scores show performance distance to respective base approach. }
    \label{tab:valnov_EGgraphs_ablation}
\end{table}

\section{Example CCKGs}
\begin{figure*}
    \centering
    \begin{subfigure}[b]{0.45\textwidth}
         \centering
         \includegraphics[width=\textwidth]{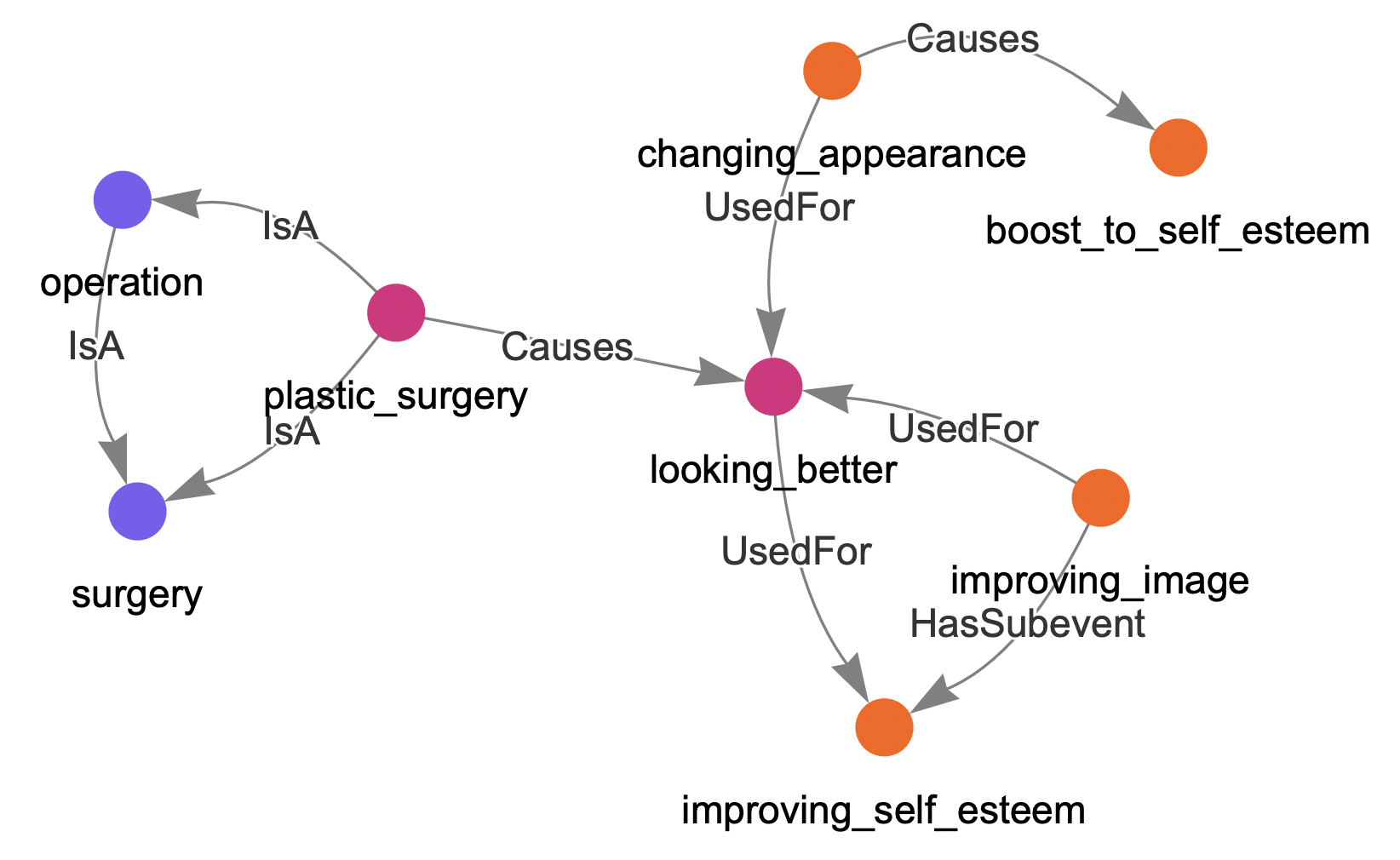}
         \caption{Premise:~\textit{Cosmetic surgery makes people feel whole again.} \\Conclusion:~\textit{Cosmetic surgery improves self esteem.}}
     \end{subfigure}
     \hfill
     \begin{subfigure}[b]{0.45\textwidth}
         \centering
         \includegraphics[width=\textwidth]{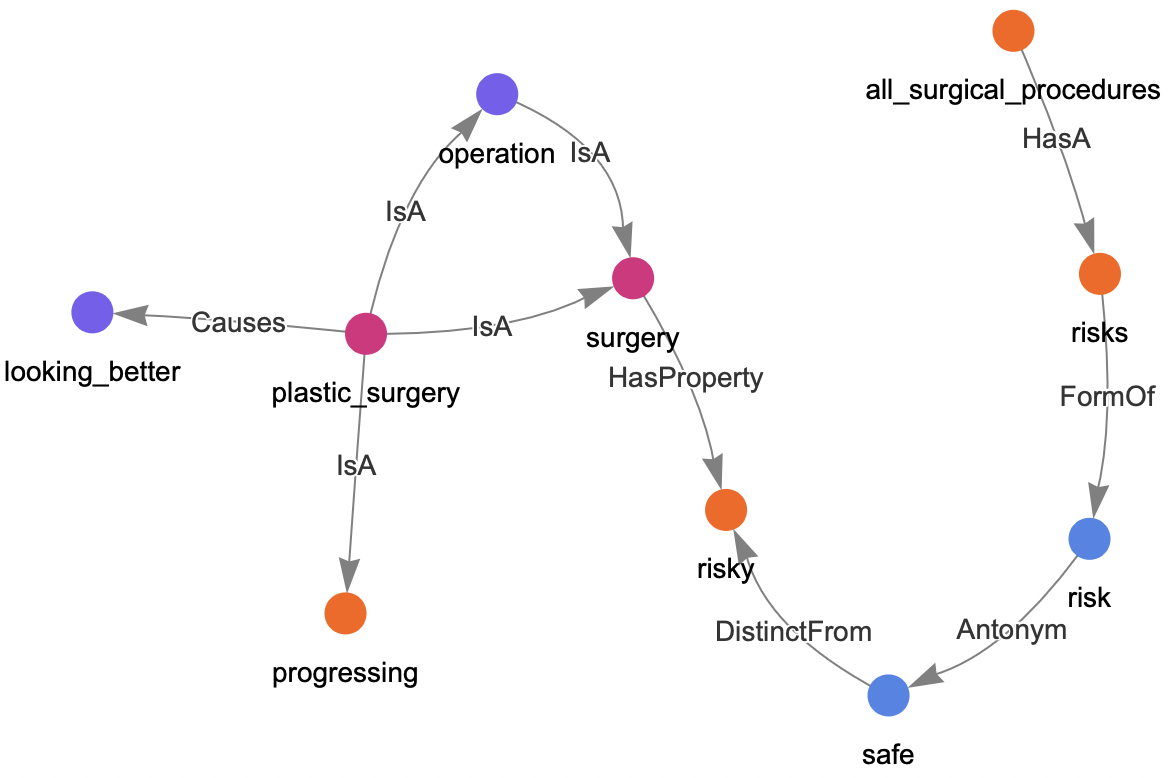} 
         \caption{Premise:~\textit{Cosmetic surgery can cause defects.} \\Conclusion:~\textit{Cosmetic surgery can be dangerous.}}
     \end{subfigure}
    \\
    \begin{subfigure}[b]{0.9\textwidth}
         \centering
         \includegraphics[width=\textwidth]{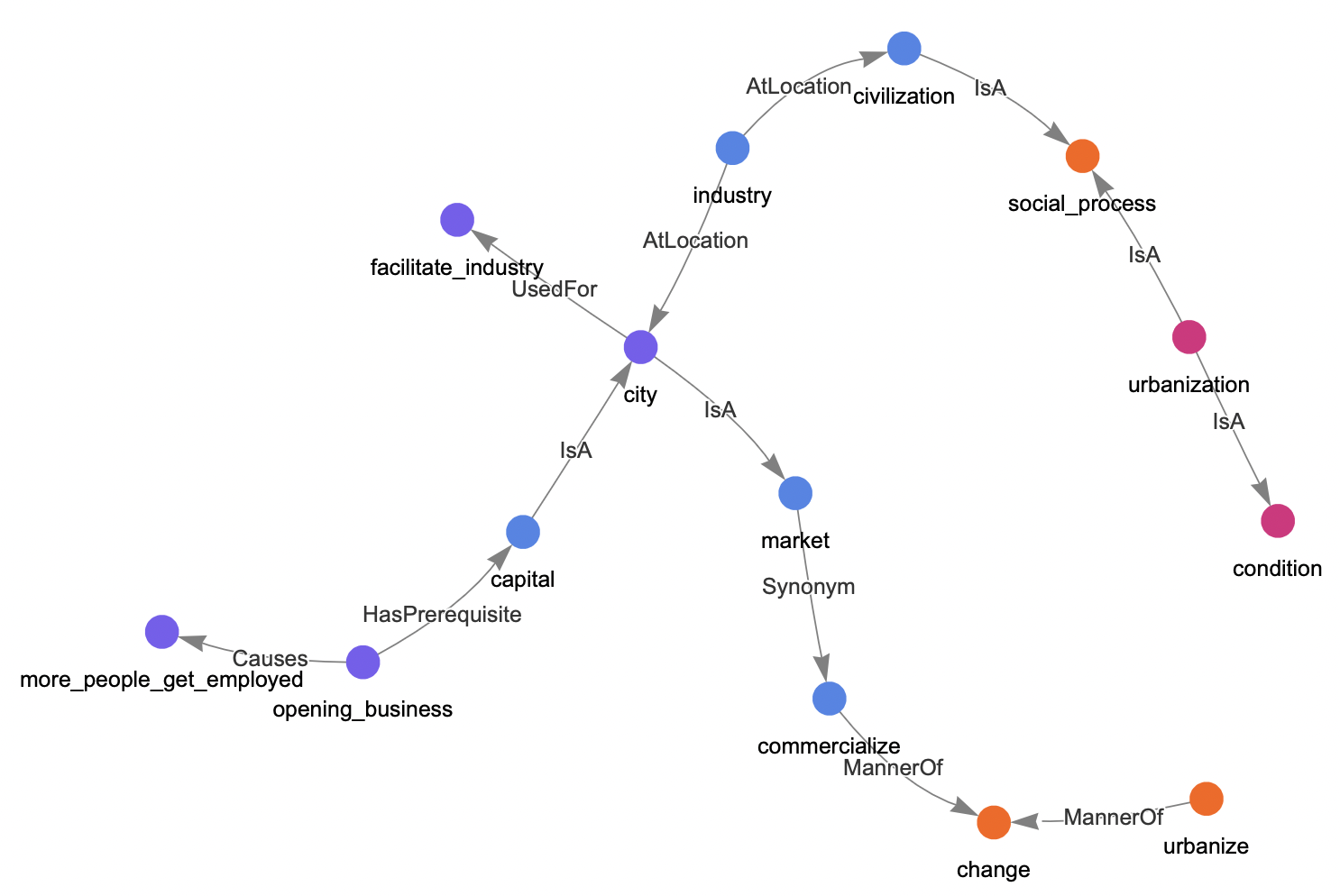}
         \caption{Premise:~\textit{Urbanization increases employment for many.} Conclusion:~\textit{Urbanization is a positive for society.}}
         \label{sub:fig:example_CCKG_1:disambigous}
     \end{subfigure}
    
    \caption{Example CCKGs for arguments from ExplaGraphs dev set. Graphs are pruned CCKGs extracted from CN without $\mathrm{RelatedTo}$ with $m=3$. \F\ref{sub:fig:example_CCKG_1:disambigous} has the disambiguity problem: $\mathrm{capital}$ is once used as city, and once as financial asset.}
    \label{fig:example_CCKG_1}
\end{figure*}

\begin{figure*}
    \centering
    \begin{subfigure}[b]{0.45\textwidth}
         \centering
         \includegraphics[width=\textwidth]{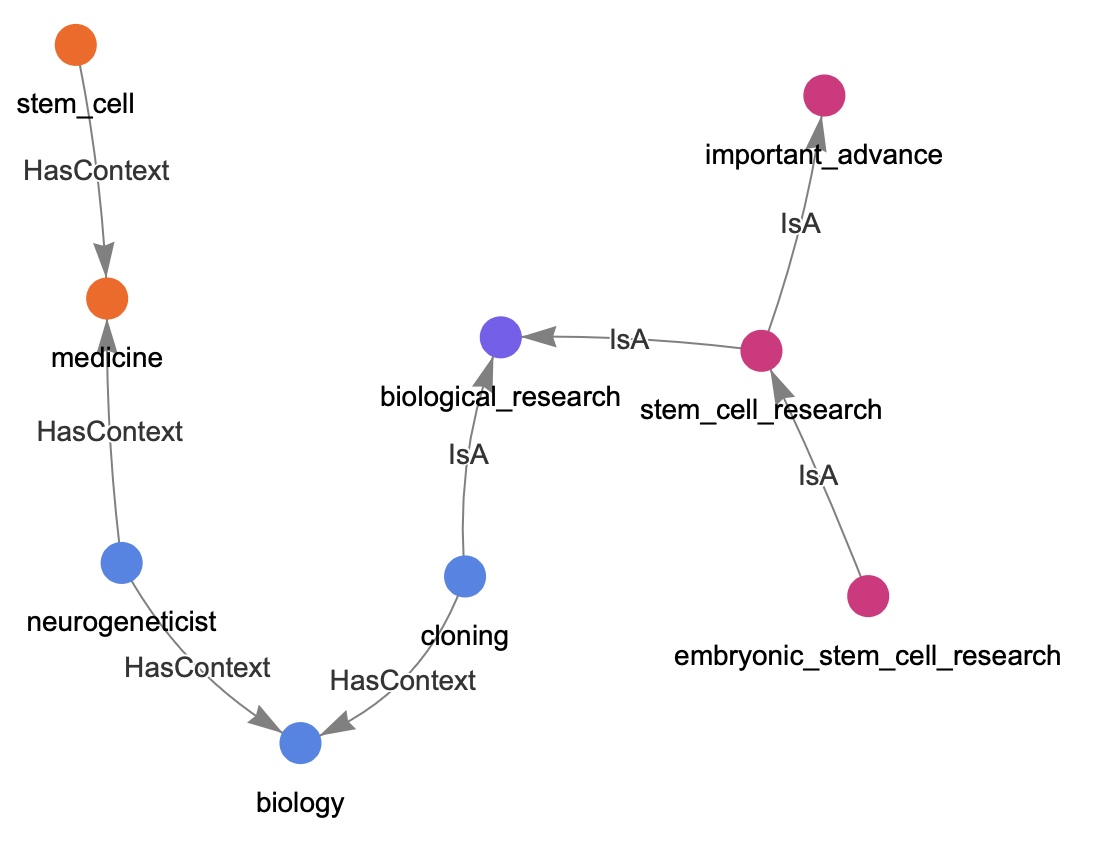}
         \caption{Premise:~\textit{Embryonic stem cell research is a no brainer.} \\Conclusion:~\textit{Embryonic stem cell research is very important to medicine.}}
     \end{subfigure}
     \hfill
     \begin{subfigure}[b]{0.45\textwidth}
         \centering
         \includegraphics[width=\textwidth]{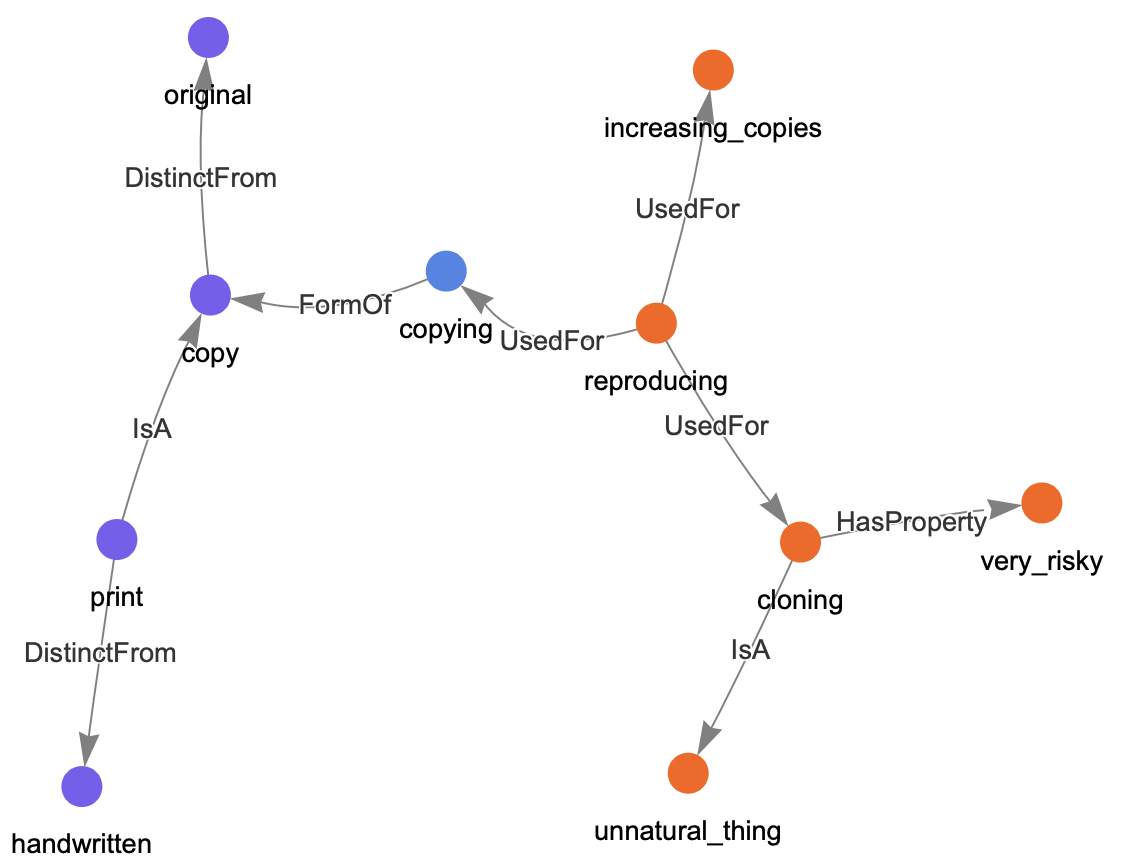} 
         \caption{Premise:~\textit{Getting your original out of the copier and putting it against the copy always shows differences.} \\Conclusion:~\textit{Cloning is inherently decreasing quality.}}
     \end{subfigure}
    \\
    \begin{subfigure}[b]{0.9\textwidth}
         \centering
        \includegraphics[width=\textwidth]{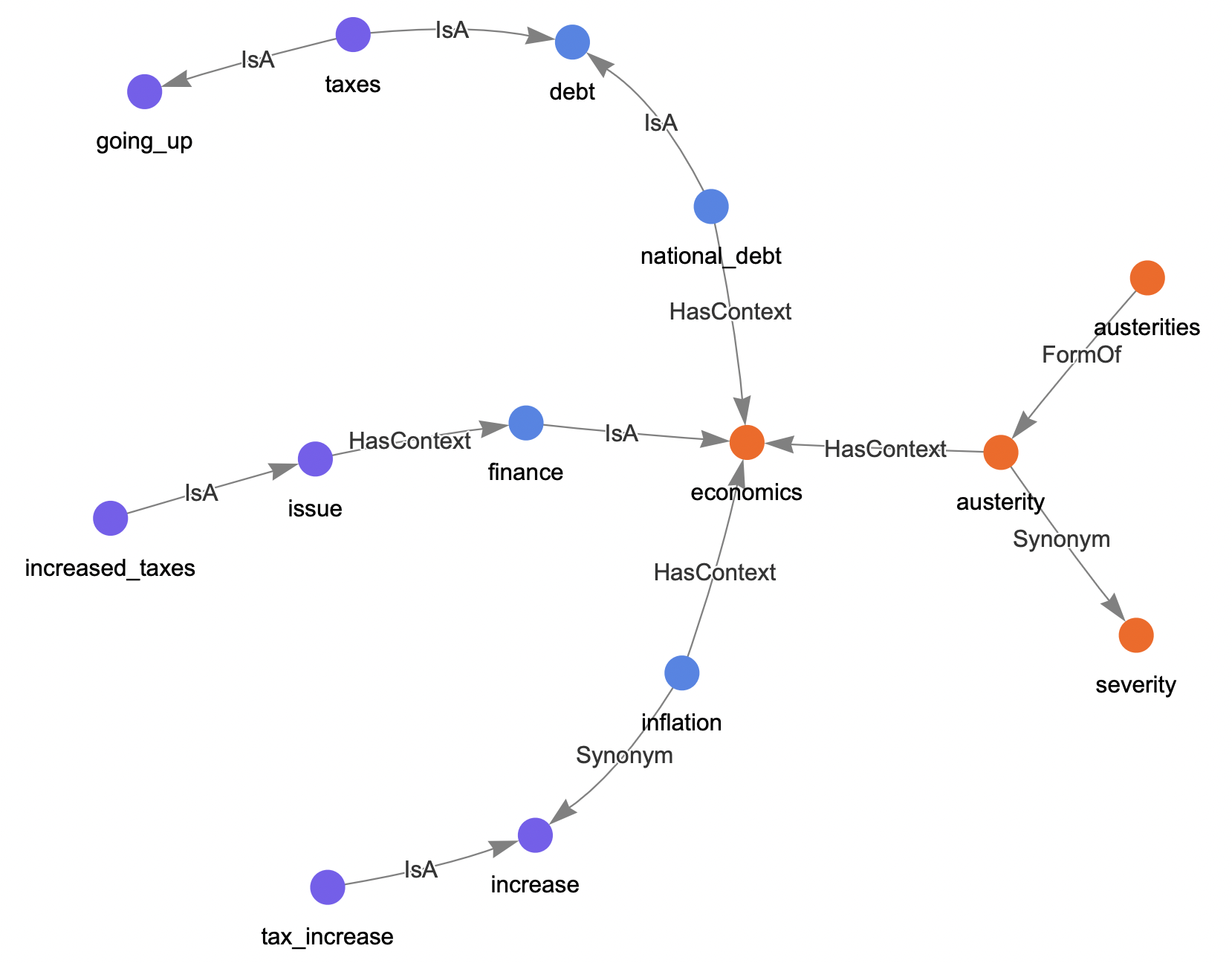} 
         \caption{Premise:~\textit{Austerity raises taxes on citizens.} \\Conclusion:~\textit{Austerity would cripple the population.}}
     \end{subfigure}
    
    \caption{Randomly selected example CCKGs for arguments from ExplaGraphs dev set. Graphs are pruned CCKGs extracted from CN without $\mathrm{RelatedTo}$ with $m=3$.}
    \label{fig:example_CCKG_2}
\end{figure*}

\begin{figure*}
    \centering
    \begin{subfigure}[b]{0.45\textwidth}
         \centering
         \includegraphics[width=\textwidth]{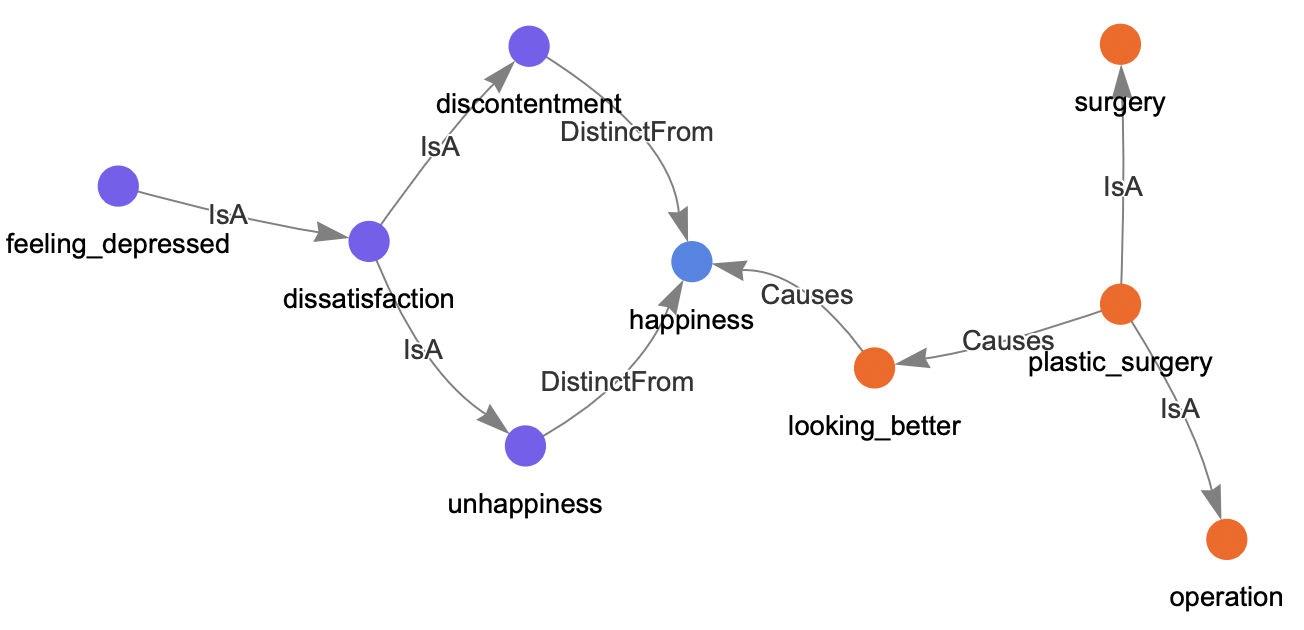}
         \caption{$m=3$, without $\mathrm{RelatedTo}$, without constituent parser}
     \end{subfigure}
     \hfill
     \begin{subfigure}[b]{0.45\textwidth}
         \centering
         \includegraphics[width=\textwidth]{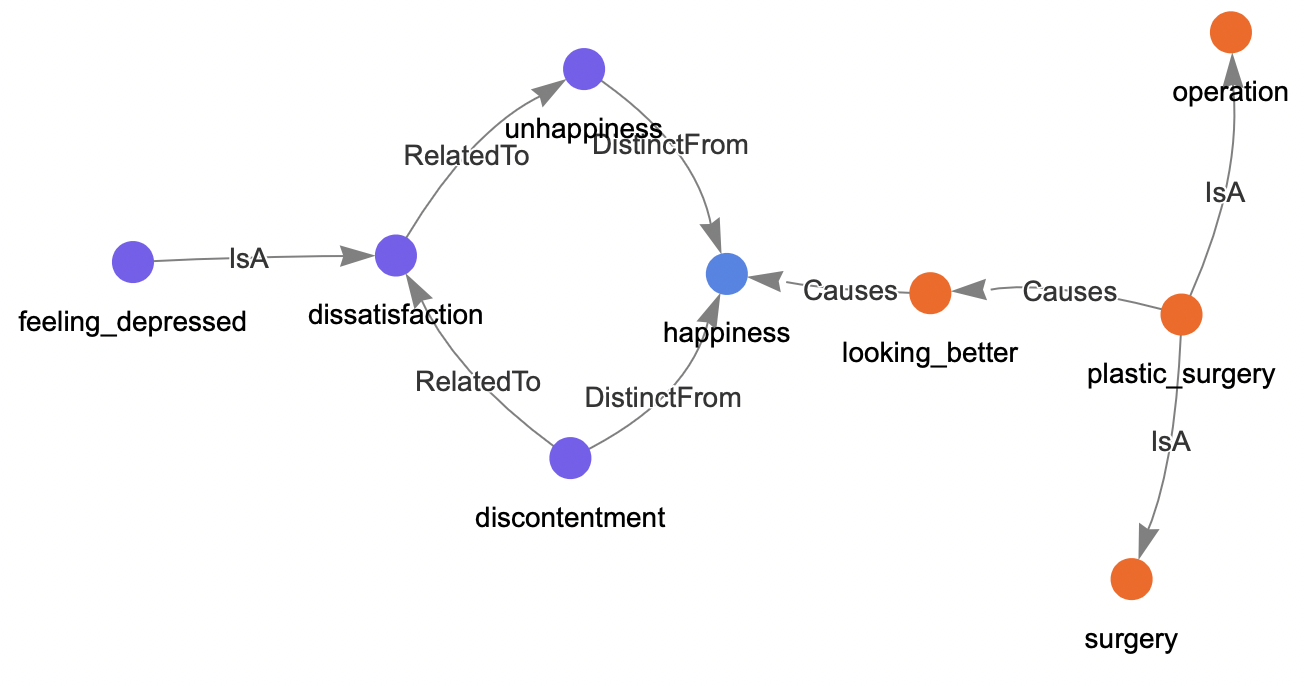}
         \caption{$m=3$, with $\mathrm{RelatedTo}$, without constituent parser}
     \end{subfigure}
    \\
    \begin{subfigure}[b]{0.9\textwidth}
         \centering
        \includegraphics[width=\textwidth]{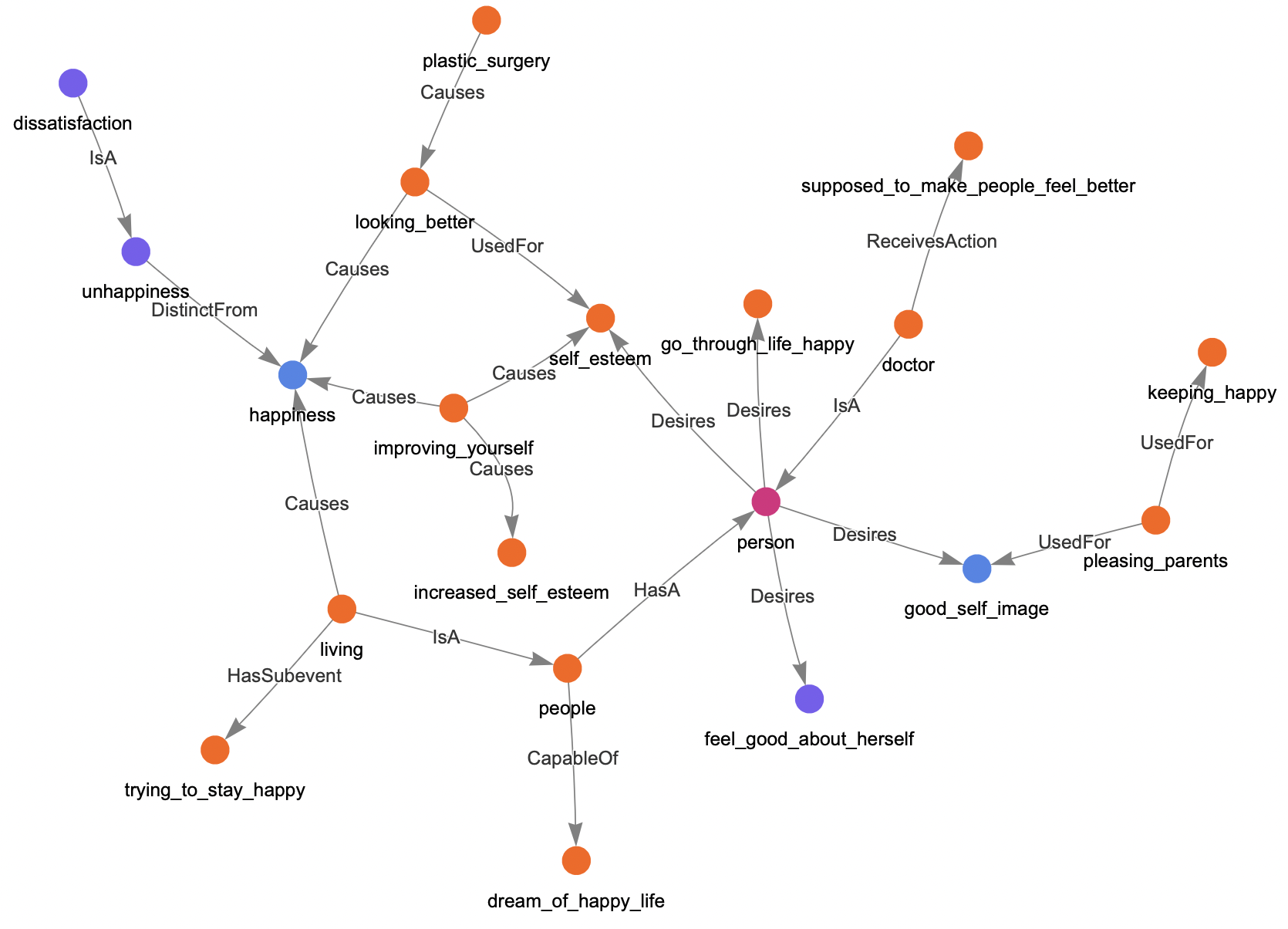} 
         \caption{$m=1$, without $\mathrm{RelatedTo}$, with constituent parser}
     \end{subfigure}
    
    \caption{Example CCKGs for premise "\textit{A person is unhappy if she is dissatisfied with her body.}" and conclusion "\textit{Plastic surgery raises patients' self esteem and allows them to lead normal happy lives.}"}
    \label{fig:example_CCKG_3}
\end{figure*}

The graphs in Figures \ref{fig:example_CCKG_1}, \ref{fig:example_CCKG_2} and \ref{fig:example_CCKG_3} show extracted premise concepts in violet, conclusion concepts in orange and intermediate concepts in blue. Concepts which are extracted for both, the premise and the conclusion, are shown in pink. 
Visualizations were done with \textit{pyvis} \\(\url{www.github.com/WestHealth/pyvis}).

\end{document}